%% file: main.tex
\pdfoutput=1
\documentclass{article}

\usepackage{microtype}
\usepackage{graphicx}
\usepackage{booktabs} %

\usepackage{hyperref}

\usepackage[accepted]{icml2024}

\usepackage{amsmath}
\usepackage{amssymb}
\usepackage{mathtools}
\usepackage{amsthm}

\usepackage[capitalize,noabbrev]{cleveref}
\usepackage{subcaption}
\usepackage{tcolorbox}
\usepackage{wrapfig}
\usepackage{multirow}

\usepackage{bbding}
\usepackage{pifont}
\usepackage[nointegrals]{wasysym}
\usepackage{xcolor}
\usepackage{bm}
\usepackage{textcomp}
\usepackage{enumitem}
\usepackage[normalem]{ulem}
\definecolor{airforceblue}{rgb}{0.36, 0.54, 0.66}
\newcommand{\blue}[1]{\textcolor{black}{#1}}

\input{math_commands.tex}

\usepackage{url}

\theoremstyle{plain}

\theoremstyle{definition}

\theoremstyle{remark}

\usepackage[textsize=tiny]{todonotes}

\icmltitlerunning{Differentially Private Representation Learning via Image Captioning}

\definecolor{ao}{rgb}{0.0, 0.5, 0.0}
\definecolor{upsdellred}{rgb}{0.68, 0.09, 0.13}
\definecolor{myblue}{RGB}{100,149,237}

\newcommand{\outline}[1]{}

\newcommand{\revisioncolor}[1]{#1}

\begin{document}

\twocolumn[
\icmltitle{Differentially Private Representation Learning via Image Captioning}

\icmlsetsymbol{equal}{*}
{\fontsize{9.5pt}{11pt}\selectfont
\begin{icmlauthorlist}
\icmlauthor{Tom Sander}{equal,meta,poly}
\icmlauthor{Yaodong Yu}{equal,meta,berkeley}
\icmlauthor{Maziar Sanjabi}{meta}
\icmlauthor{Alain Durmus}{poly}
\icmlauthor{Yi Ma}{berkeley}
\icmlauthor{Kamalika Chaudhuri}{meta,ucsd}
\icmlauthor{Chuan Guo}{meta}
\end{icmlauthorlist}
}
\icmlaffiliation{meta}{Meta}
\icmlaffiliation{poly}{CMAP, École polytechnique}
\icmlaffiliation{berkeley}{UC Berkeley}
\icmlaffiliation{ucsd}{UCSD}
\icmlcorrespondingauthor{Tom Sander}{tomsander@meta.com}
\icmlcorrespondingauthor{Yaodong Yu}{yyu@eecs.berkeley.edu}
\icmlcorrespondingauthor{Chuan Guo}{chuanguo@meta.com}
\icmlkeywords{Machine Learning, ICML}
\vskip 0.3in
]

\printAffiliationsAndNotice{\icmlEqualContribution} %

\begin{abstract}
Differentially private (DP) machine learning is considered the gold-standard solution for training a model from sensitive data while still preserving privacy. However, a major barrier to achieving this ideal is its sub-optimal privacy-accuracy trade-off, which is particularly visible in DP representation learning. Specifically, it has been shown that under modest privacy budgets, most models learn representations that are not significantly better than hand-crafted features. In this work, we show that effective DP representation learning can be done via image captioning and scaling up to internet-scale multimodal datasets. Through a series of engineering tricks, we successfully train a DP image captioner (DP-Cap) on a 233M subset of LAION-2B \emph{from scratch} using a reasonable amount of computation, and obtaining unprecedented high-quality image features that can be used in a variety of downstream vision and vision-language tasks. 
For example, under a privacy budget of $\varepsilon=8$ \revisioncolor{for the LAION dataset}, a  linear classifier trained on top of learned DP-Cap features attains $65.8\%$ accuracy on ImageNet-1K, considerably improving the previous SOTA of $56.5\%$.
Our work challenges the prevailing sentiment that high-utility DP representation learning cannot be achieved by training from scratch.
Code is available at \url{https://github.com/facebookresearch/dpcap}.
\end{abstract}

\input{1-intro}

\input{2-prelim}

\input{3-method}

\input{4-experiments}

\input{5-discussion}

\vspace{-0.15cm}
\revisioncolor{
\section*{Acknowledgements}
This research was partially supported as a BAIR Open Research Common Project with Meta. We thank Quentin Garrido for his time and insights regarding the significance of vision backbones quality evaluations.
}

\vspace{-0.15cm}
\section*{Impact Statement}
Our work focuses on developing privacy-preserving techniques for machine learning. Progress in this direction has
significant potential for evolving the concept of privacy for
AI. The purpose of our work is to push the boundary of what is achievable for representation learning through differentially private mechanisms, towards determining the
technical feasibility of applying DP to tackle real-world AI
privacy challenges.

\bibliographystyle{icml2024}
\bibliography{paper}

\newpage
\appendix
\onecolumn
\input{sec_appendix}

\end{document}

%% file: math_commands.tex
\usepackage{amsmath,amsfonts,bm}

\def\eqref#1{equation~\ref{#1}}

\def\1{\bm{1}}

\newcommand{\calM}{\mathcal{M}}
\newcommand{\gauss}{\mathcal{N}}

\DeclareMathAlphabet{\mathsfit}{\encodingdefault}{\sfdefault}{m}{sl}
\SetMathAlphabet{\mathsfit}{bold}{\encodingdefault}{\sfdefault}{bx}{n}

%% file: 1-intro.tex
\begin{figure*}[t!]
     \centering
     \includegraphics[width=0.99\textwidth]{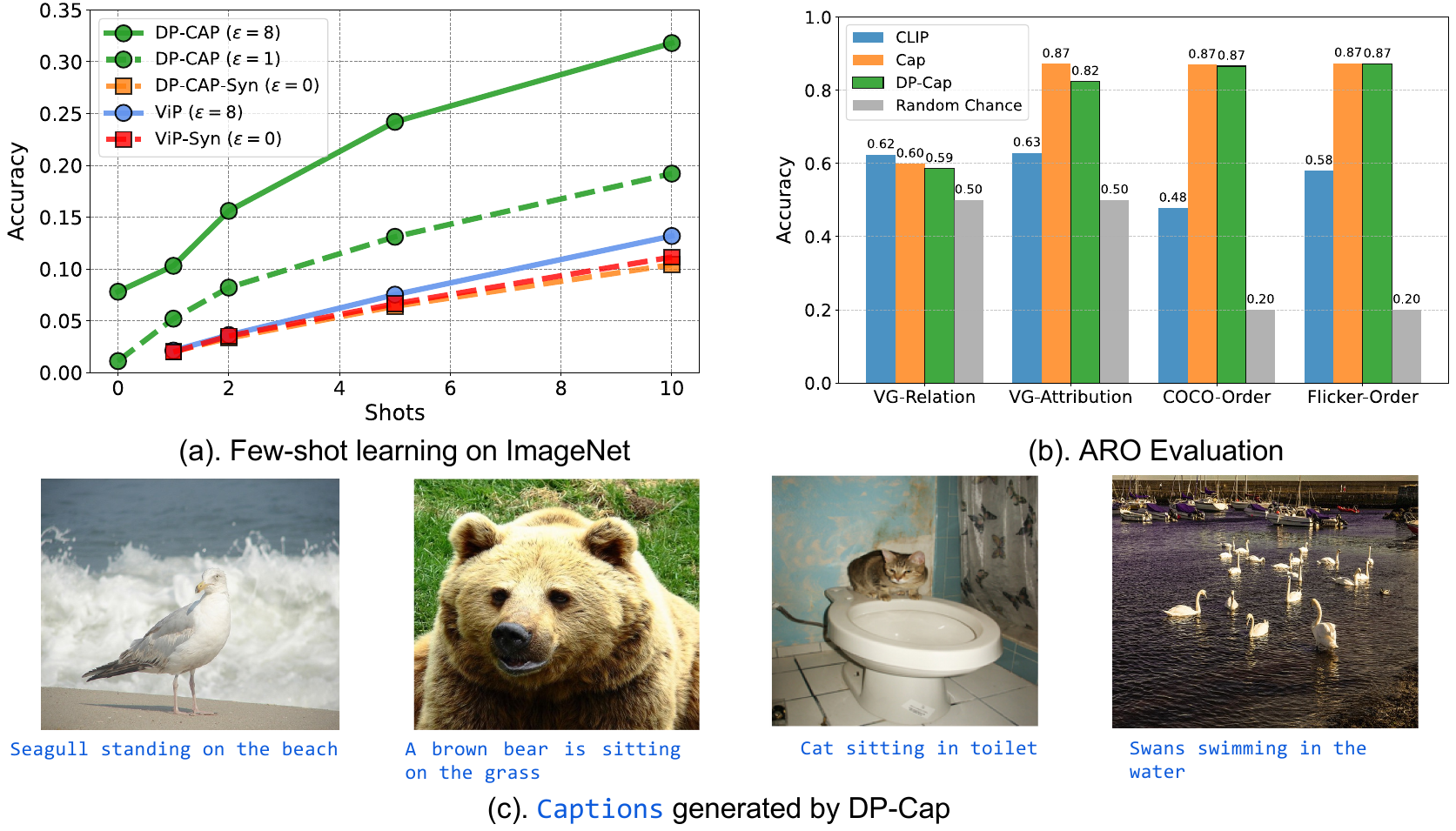}
     \vspace{-0.05in}
    \caption{(a) Few-shot ImageNet-1K linear probe accuracy comparison between DP-Cap (ours) and ViP~\citep{yu2023vip} (previous SOTA). DP-Cap learns better image representations using the same training data and privacy budget, and considerably surpasses synthetic initialization (\emph{syn}). \revisioncolor{The privacy budget $\varepsilon$ is for the LAION dataset, and the linear classifiers are trained without DP.} (b) Compositional understanding evaluation on the ARO benchmark~\citep{yuksekgonul2022and}. DP-Cap performance is close to non-private Cap and outperforms non-private CLIP. (c) Captions generated by DP-Cap on images from the MS-COCO 2017~\citep{lin2015microsoft} test set.}
    \label{fig:1}
    \vspace{-0.1in}
\end{figure*}

\section{Introduction}
\label{sec:introduction}

Differentially private (DP; \citet{dwork2006calibrating}) model training is an effective strategy for privacy-preserving ML on sensitive data. For most optimization-based learning algorithms, DP-SGD~\citep{song2013stochastic, abadi2016deep} can be readily applied to obtain models with rigorous DP guarantee. Regrettably, DP training has also been marred by a sub-optimal privacy-utility trade-off, with model utility severely lagging behind their non-private counterpart~\citep{jayaraman2019evaluating, tramer2020differentially, kurakin2022toward}. At the core of this unfavorable trade-off is the \emph{difficulty of DP representation learning}. \citet{tramer2020differentially} showed that when DP training from scratch under a low-to-moderate privacy budget, most models do not learn useful representations, with the quality of learned representations worse than even handcrafted features. 
These observations naturally lead to the research question: \emph{``How does one learn useful representations with DP training?''}

One plausible reason for the failure of prior attempts at DP representation learning is the lack of training data. Indeed, DP limits the information content of each training sample via the privacy budget $\varepsilon$, inducing a privacy-accuracy-sample size tradeoff; thus a substantially larger training dataset is required to extract the same amount of information to train the model. As the vast majority of prior work only utilize small to moderate scale classification datasets such as CIFAR-10~\citep{krizhevsky2009learning} and ImageNet~\citep{deng2009imagenet}, the amount of training data is simply insufficient for learning high-quality representations under DP~\citep{tramer2020differentially}.
\citet{yu2023vip} made partial progress towards this through \emph{self-supervised learning} (SSL) on \emph{internet-scale data}. By training a masked autoencoder (MAE; \citet{he2022masked}) using DP-SGD on a 233M subset of the LAION-2B dataset~\citep{Schuhmann2022LAION5BAO}, the model learned image representations that are on-par with non-private AlexNet~\citep{krizhevsky2012imagenet} trained on ImageNet---the first deep learning model to outperform handcrafted features and a major cornerstone for representation learning.
However, the MAE objective also promotes the model to learn extraneous details in the image that may not be helpful for obtaining generalizable representations, severely limiting the potential of this approach for DP.

We adopt a different approach of DP training via image captioning on internet-scale multimodal datasets. The reason is twofold: \textbf{1.} Text caption provides a concise summary of the training image and serves as better supervision compared to image-only SSL~\citep{tschannen2023image}. Under the constraint on information content from DP, we hypothesize that it provides substantially more efficient information extraction under DP training.
\textbf{2.} Image captioning is well-aligned with the prerequisites of DP-SGD such as having an instance-separable loss.
We apply this method on a 233M subset of LAION-2B to train a DP image captioning model (DP-Cap), whose learned representations surpass previous SOTA---ViP~\citep{yu2023vip}---by a large margin. As depicted in Figure~\ref{fig:1}(a), our model trained with a privacy budget of $\varepsilon=8$ shows substantial improvements on
downstream tasks compared to ViP, both trained on the same dataset. To achieve this, we also made crucial improvements to the efficiency of the DP training pipeline, \textbf{reducing the compute cost by close to $\mathbf{5\times}$} on the largest model.

The image representations learned by DP-Cap also exhibit strong performance for multimodal tasks that require alignment of image and text features,
the first occurrence for models trained from scratch with DP; see Figure~\ref{fig:1}(b).
As a qualitative evaluation, we also use the trained DP-Cap model to caption several images from the MS-COCO 2017~\citep{lin2015microsoft} test set in Figure~\ref{fig:1}(c) and Appendix~\ref{subsec:appendix-image-caption}. The resulting captions are grammatically correct and semantically coherent, while (close to) accurately describing contents of the image; this is interesting because our model has only been exposed to language supervision from LAION, which are far from being flawless.
Our results suggest that DP training on internet-scale multimodal datasets can be a viable approach for obtaining high-utility learned representations.

%% file: 2-prelim.tex
\section{Background and Related Work}
\label{sec:background}

\textbf{Vision-language pre-training.} Many modern ML datasets such as Conceptual Captions~\citep{changpinyo2021conceptual}, LAION~\citep{schuhmann2021laion} and DataComp~\citep{gadre2023datacomp} consist of aligned image-text pairs where the image and text contain roughly similar semantic information. One can leverage the aligned nature of the training data to pre-train \emph{vision-language models} (VLMs) that connect the two modalities, whose representations perform more general multi-modal tasks. Contrastive learning-based techniques such as CLIP~\citep{radford2021learning} and BLIP~\citep{li2022blip} are also applicable for pre-training VLMs.
Doing so not only learns high-quality image and text representations but also introduces new multi-modal capabilities such as cross-modal retrieval and zero-shot prediction~\citep{radford2021learning}.
Recent work by \cite{tschannen2023image} shows an image captioning approach (predicting text captions from images) is a viable alternative to contrastive learning and can lead to models with robust performance.

\textbf{Differential privacy \citep{dwork2006calibrating}.} In the following, we denote by $\calM$ a randomized learning algorithm, which takes a dataset $\mathcal{D}$ containing $N$ samples and produces a machine learning model $\bm{\theta}$ through the process $\calM(\mathcal{D})$.
A randomized mechanism $\calM$ is $(\varepsilon, \delta)$-DP if, for any two adjacent datasets $\mathcal{D}$ and $\mathcal{D}^{\prime}$ differing by a single sample, and for any subset $\mathcal{O}\subset \mathbf{Im}(\calM)$:
\begin{equation}\label{eq:def-dp}
    \mathbf{P}[\calM(\mathcal{D}) \in \mathcal{O}] \leq \mathbf{P}[\calM(\mathcal{D}^{\prime}) \in \mathcal{O}] \exp( \varepsilon) + \delta.
\end{equation}
We adopt the leave-one-out notion of adjacency in this work, \emph{i.e.}, $\mathcal{D} = \mathcal{D}^{\prime} \cup \{\mathbf{x}\}$ for some sample $\mathbf{x}$ or vice versa. DP bounds the extent to which any potential adversary can infer information about the dataset $\mathcal{D}$ after observing the algorithm's output.
In the context of ML, this implies that if we obtain the model $\bm{\theta}$ through a DP training algorithm $\calM$ then its training data is provably difficult to recover or infer~\citep{balle2022reconstructing, guo2022bounding,guo2023analyzing}.

\textbf{DP-SGD}~\citep{song2013stochastic, abadi2016deep} is predominant differentially private algorithm for training deep neural networks (DNNs).
At each gradient step $k$, a batch $\mathcal{B}_k$ is sampled where each example from the training data is chosen randomly with probability $q=B/N$, where $B$ represents the average batch size. 
For $C >0$, define the clipping function for any $X \in \mathbb{R}^d$ by $\mathrm{clip}_C(X) = C\cdot X/\Vert X \Vert$ if $\Vert X \Vert \geq C$ and $\mathrm{clip}_C(X) = X$ otherwise.
Given model parameters $\bm{\theta}_k$, DP-SGD defines the update
$\bm{\theta}_{k+1} = \bm{\theta}_k -\eta_k \widetilde{\mathbf{g}}_k$ where $\eta_k$ is the step size and $\widetilde{\mathbf{g}}_k$ is given by:
\begin{equation}
\label{eq:DP_SGD}
{\small 
\widetilde{\mathbf{g}}_k := \frac{1}{B} \left[ \sum_{i \in \mathcal{B}_k} \text{clip}_C\left(\nabla_{\bm{\theta}} \ell_i(\bm{\theta}_{k})\right) + \gauss \left(0, C^2 \sigma^2 \mathbf{I} \right) \right],}
\end{equation}
where $\ell_i(\bm{\theta})$ is the  per-sample loss function evaluated at sample $\mathbf{x}_i$. 
We also use the term ``DP-SGD'' loosely to refer to the category of gradient-based optimization algorithms that operate on the noisy gradient, \emph{e.g.}, Adam~\citep{kingma2014adam}.
The privacy analysis of DP-SGD relies on composition of multiple steps.
One particularly powerful analysis framework amenable to such compositions relies on a variant of DP called R\'{e}nyi differential privacy (RDP) \citep{mironov2017renyi}.
An advantage of RDP is its additive composition property, where the privacy guarantees of a sequence of mechanisms can be combined with amplification by subsampling~\citep{wang2019subsampled} and then translated to $(\varepsilon, \delta)$-DP~\citep{balle2020hypothesis, gopi2021numerical}. 
In this work, we adopt this accounting technique.

\textbf{Scaling up DP-SGD training.} DP training is a theoretically and empirically proven remedy against unintended training data memorization.
Even models with large $\varepsilon$ (\emph{e.g.}, $\varepsilon=100$) can empirically defend against privacy attacks~\citep{carlini2021membership, guo2023analyzing}. 
Despite its great appeal, DP training also carries a significant drawback of large drop in model utility~\citep{abadi2016deep,tramer2020differentially}.
For example, the SOTA performance on ImageNet when training from scratch with a DP guarantee of $\varepsilon=8$ is 39.2\%~\citep{sander2023tan}; in comparison, the non-private performance on ImageNet when training from scratch can reach 88\%~\citep{touvron2022deit} or higher. This degradation in model utility also translates to poorly learned representations, as \citet{tramer2020differentially} showed that even handcrafted features can rival ones learned through DP training.

\citet{yu2023vip} made the first step towards obtaining high-utility learned representations through scaling DP training. They proposed self-supervised learning (SSL) on internet-scale data as a solution for the privacy-utility trade-off in DP representation learning.
Among the numerous SSL algorithms, the authors observed that the reconstruction-based approach of masked autoencoder (MAE; \citet{he2022masked}) is compatible with the requirements of DP-SGD.
By leveraging weight initialization through synthetic pre-training, the authors were able to obtain high-utility learned representations at a strict privacy budget of $\varepsilon=8$. 
Compared to ViP~\citep{yu2023vip}, we demonstrate that the image captioning approach (see Section ~\ref{subsec:pretrain-captioners-with-DPSGD}) learns much better image representations by utilizing the additional text supervision.

%% file: 3-method.tex
\vspace{-0.05in}
\section{Approach
}\label{sec:approach}

We describe in detail our approach of DP representation learning via image captioning. We first argue why image captioning is intuitively a suitable objective for obtaining better image representations via DP-SGD training (section \ref{subsec:pretrain-captioners-with-DPSGD}). Then, we elucidate the technical challenges that we resolved to make DP training viable and effective for image captioning (section \ref{subsec:effective_strategies_DPCAP}).

\vspace{-0.02in}
\subsection{DP Representation Learning via Image Captioning}\label{subsec:pretrain-captioners-with-DPSGD}

\textbf{Why is vision-language pre-training suitable?} %
Given image-text aligned datasets, prior works~\citep{radford2021learning, li2022blip, tschannen2023image} showed that pre-training using language supervision is an appealing option for non-private representation learning. We hypothesize that this is true for DP representation learning as well. Compared to image-only supervision, language supervision contains a more condensed summary of the image content, allowing the model to ignore irrelevant details such as background and focus on objects of interest and their relationships. 
This is especially helpful for DP since the model needs to extract as much useful information as possible from each sample given the privacy budget $\varepsilon$.
Captioning could thus enhance the privacy-utility-sample size trade-off in DP, considering it requires less information per sample.

In addition, we show that vision-language pre-training supports a very large batch size, much larger than what is typically used in image-only pre-training~\citep{radford2021learning, li2022blip, yu2022coca}. 
This subtle aspect is in fact crucial for reducing the effective noise in DP-SGD~\citep{li2021large}, which allows the model parameters to converge to a stable solution with lower training loss (see Section~\ref{subsec:effective_strategies_DPCAP}).

\textbf{Vision-language pre-training via image captioning.} 
Perhaps the most popular approach for vision-language pre-training is contrastive language image pre-training (CLIP; \citet{radford2021learning}) as well as its variants~\citep{mu2022slip, li2023scaling}.   
However, the contrastive loss used in these methods is not an additive function over the samples, \emph{i.e.}, it cannot be written in the form $\sum_{i} \ell_i$, where $\ell_i$ depends only on the $i$-th sample.
Thus, DP-SGD~(\emph{cf.} \eqref{eq:DP_SGD}) cannot be directly applied.

Unlike contrastive learning, the image captioning approach~\citep{sariyildiz2020learning,desai2021virtex, tschannen2023image} aligns well with DP-SGD training.
Specifically, an image captioner is trained to predict captions based on their corresponding images. 
The training objective of the image captioner for one image-text pair $[\mathbf{x}^{\text{img}}, \mathbf{z}^{\text{text}}]$ is to minimize over $\bm{\theta}:=\{\bm{\theta}_{\textsf{enc}}, \bm{\theta}_{\textsf{dec}}\}$ the following loss:
\vspace{-0.1in}
\revisioncolor{
\begin{align}\label{eq:loss-image-captioner}
 &L_{\text{Cap}}(\bm{\theta}) 
 := \frac{1}{T}\sum_{t=0}^{T-1} \ell_{\text{CE}}\Big({z}^{\text{text}}_{t+1}, \varphi\big(\mathbf{z}^{\mathrm{img}}, \mathbf{z}_{1:t}^{\mathrm{text}}; \bm{\theta}_{\textsf{dec}}\big)\Big), \\
 &\mathrm{where}\, \mathbf{z}^{\mathrm{img}}=\underbrace{\psi(\mathbf{x}^{\text{img}}; \bm{\theta}_{\textsf{enc}})}_{\text{image embedding}}, \,\, \mathbf{z}^{\mathrm{text}}_{1:t} = \underbrace{{z}^{\text{text}}_1,
\ldots, {z}^{\text{text}}_{t}}_{\text{first $t$ tokens}}. \nonumber 
\vspace{-0.2in}
\end{align}
}

\vspace{-0.1in}
\noindent
We use $\mathbf{z}^{\text{text}}$ denotes the caption token sequence $ \{{z}^{\text{text}}_1,
\ldots, {z}^{\text{text}}_{T}\}$, \revisioncolor{ and we let $\bm{z}_{1:0}^{\mathrm{text}} = \emptyset$.} 
The image captioner consists of two parts: the image encoder $\psi(\cdot; \bm{\theta}_{\textsf{enc}})$ and the text decoder $\varphi(\cdot; \bm{\theta}_{\textsf{dec}})$. 
The rationale behind the design of \eqref{eq:loss-image-captioner} is that the image encoder maps the input image $\mathbf{x}^{\text{img}}$ to an embedding vector, and the text decoder takes the image embedding $\psi(\mathbf{x}^{\text{img}}; \bm{\theta}_{\textsf{enc}})$ and the first $t$ caption tokens $\{{z}^{\text{text}}_1,
\ldots, {z}^{\text{text}}_{t}\}$ as inputs and predicts the next caption token ${z}_{t+1}^{\text{text}}$.
Both the encoder and decoder are trained to maximize the log-likelihood of the correct next token. 
Equation~\ref{eq:loss-image-captioner} corresponds to the loss function for an image-text pair $[\mathbf{x}^{\text{img}}, \mathbf{z}^{\text{text}}]$; summing over all the samples in a batch gives the complete empirical loss in an additive form, which is directly compatible with DP-SGD.

\revisioncolor{
\textbf{Interpretation of DP bound for image-text pairs.} $\varepsilon$-DP bounds the amount of information extracted from each training sample by $\varepsilon$. For DP-Cap, each sample is made of \{image + caption\}, while ViP~\cite{yu2023vip} utilizes \{image\} only. Consequently, DP-Cap inherently offers an equivalent or better privacy guarantee for each image. 
To better understand this, we can consider membership inference attacks~\citep{shokri2017membership} as an example.
Suppose $\varepsilon$-DP upper bounds the success rate of the membership inference attack (when given the image-text pair) against DP-Cap as less or equal to $p$. 
Then the MIA success rate when given only the image can be at most $p$ since the attacker has strictly less information. 
This is exactly the upper bound for the success rate of a membership inference attack against ViP. }

\vspace{-0.2cm}
\subsection{Strategy for Effective DP Training}\label{subsec:effective_strategies_DPCAP}

\begin{figure}[t!]
  \centering
  \includegraphics[width=\linewidth]{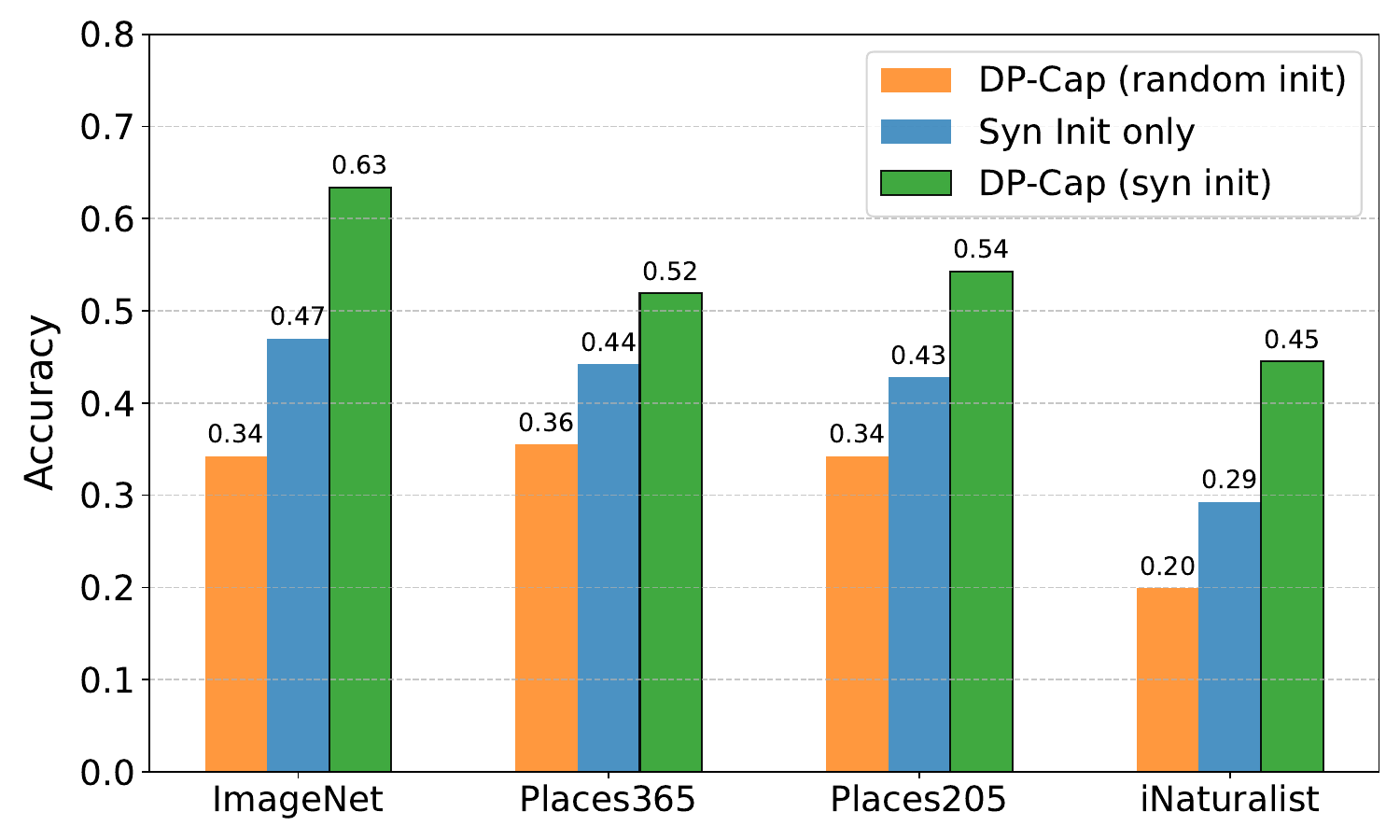}
  \vspace{-0.25in}
  \caption{Impact of synthetic initialization on DP-Cap. 
  The learned image representation benefits substantially from initializing on the Shaders21K dataset. The
  gap between DP-Cap (random init) and DP-Cap (syn init) can be as large as 24\% when evaluated using linear probing on ImageNet.}
  \vspace{-0.25in}
  \label{fig:benefit-syn-pretrain}
\end{figure}

Although image captioning has demonstrated impressive representation learning capabilities in the non-private regime, adapting it to DP training requires careful considerations.
To obtain a useful pre-trained model, one needs to train for a sufficient number of steps under a low effective noise, both of which are at odds with obtaining a strong privacy guarantee.
We detail the strategy we used to handle this trade-off when training the image captioner. 

\begin{figure*}[t!]
  \centering
  \begin{subfigure}{0.49\textwidth}
  \centering
    \includegraphics[width=\linewidth]{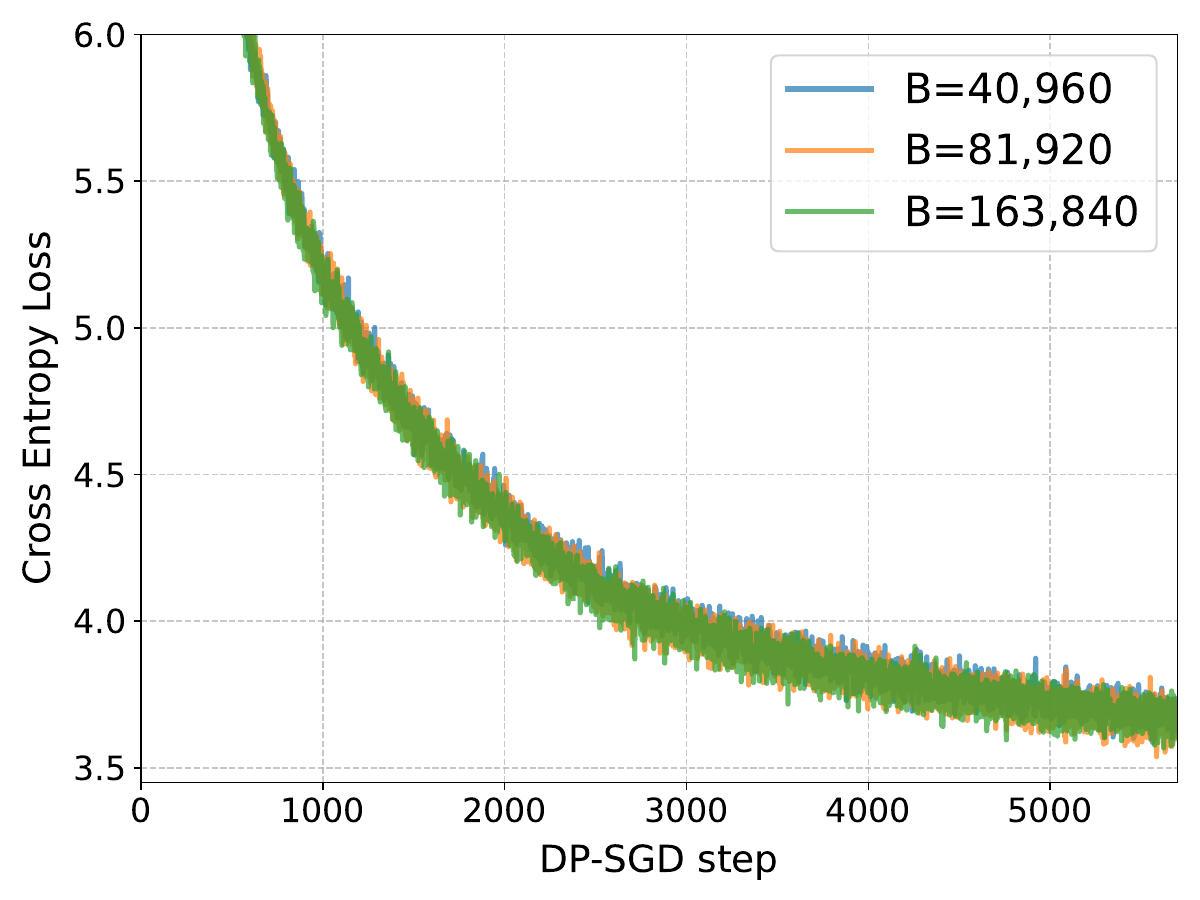}
    \caption{DP-Cap scales gracefully to extreme batch sizes}
  \end{subfigure}
  \begin{subfigure}{0.49\textwidth} 
  \centering
  \includegraphics[width=0.95\linewidth]{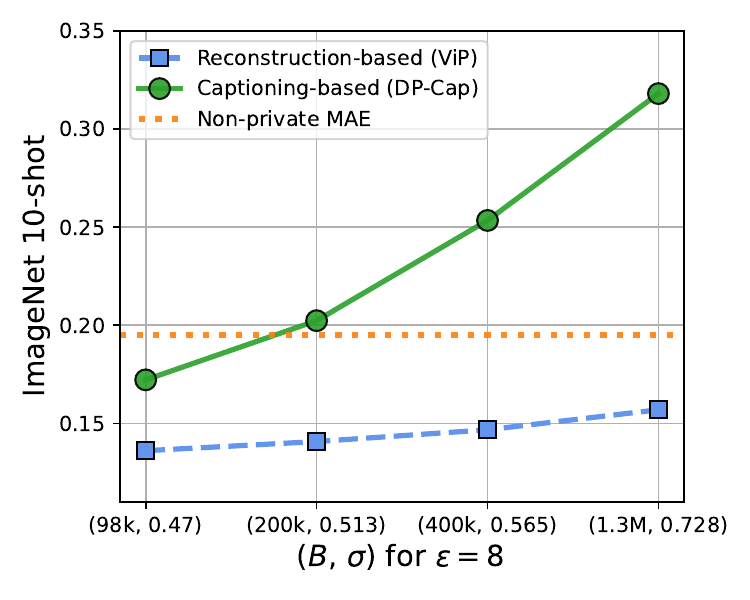}
    \caption{Performance improves consistently with larger batch size}
  \end{subfigure}
  \vspace{-0.05in}
  \caption{(a) We fix the effective noise $\sigma/B = 5.6 \times 10^{-7}$ \blue{(corresponding to our (B, $\sigma$) = (1.3M, 0.728))} and show that the loss is remarkably consistent across different batch sizes, allowing us to effectively scale up batch size to improve the SNR. (b) Performance from 4 sets of parameters that provide $\varepsilon=8$, with constant number of steps 5708. From batch size 98k (used in ViP~\citep{yu2023vip}), to our 1.3M batch size.
  In contrast to ViP, DP-Cap successfully leverages the better SNR and learns features that achieve substantially better 10-shot accuracy on ImageNet even compared to a \textbf{non-private} MAE \citep{he2022masked} trained on the same dataset (see Appendix~\ref{sec:appendix_optim}).}
  \label{fig:3}
  \vspace{-2ex}
\end{figure*}

\textbf{Sufficient number of training steps.} 
We address this challenge via synthetic pre-training. 
Image representations learned by \textit{DP-Cap (random init)}  outperform those of \textit{ViP (random init)} as evidenced in table~\ref{tab:appendix_randominit} in Appendix~\ref{sec:appendix-additional-exp-results}. 
Interestingly, ~\citet{yu2023vip} \revisioncolor{and ~\citet{tang2024differentially} have shown that synthetic images consisting of only textures can provide a good initialization for training DP models without any privacy risk. 
With this initialization in ~\citet{yu2023vip}, the reconstruction-based model }can \revisioncolor{then} focus on learning dataset-specific properties rather than low-level image properties such as edge detectors, therefore expending privacy budget in a more optimal manner.
We adapt this technique of pre-training on the Shaders21K dataset~\citep{baradad2022procedural} for initialization and observe that it is even more effective with DP-Cap compared to ViP.\footnote{\revisioncolor{For the synthetic pre-training on Shders21k in DP-Cap, we modify the training of image captioning model such that the model is trained to predict missing image patches, which is similar to MAE~\cite{he2022masked}. More details can be found in Appendix~\ref{sec:appendix_optim}.  }}
As shown in Fig.~\ref{fig:benefit-syn-pretrain}, our \textit{DP-Cap (syn init)} improves over \textit{DP-Cap (random init)} by more than 24\% on ImageNet-1k linear probing. It also improves over synthetic initialization alone (\textit{Syn-init}) by more than 14\% on ImageNet-1k linear probing, whereas the gain in ViP is smaller than 6\%.

\textbf{Using extreme batch sizes to reduce effective noise.} \citet{li2021large} first showed that increasing the batch size in DP-SGD often improves the privacy-utility trade-off. This is because the effective noise added to the average gradient has magnitude $\sigma / B$ (\emph{cf.} \eqref{eq:DP_SGD}), and that increasing $B$, rather than decreasing $\sigma$, results in better privacy guarantees according to conventional accounting techniques~\citep{bun2016concentrated, dwork2016concentrated, mironov2019r, sander2023tan}.
\revisioncolor{However, in supervised learning, increasing the batch size beyond a certain point may degrade performance in both private and non-private settings~\citep{sander2024implicit}}.
Specifically,~\citet{sander2023tan} 
observed that when training a classifier from scratch with DP-SGD on ImageNet, at a fixed number of steps $S$ and fixed effective noise $\sigma/B$, the performance decreases significantly when the batch size becomes too large: for $B \in [128, 16384]$, a drop of 10\% in top-1 accuracy was observed.

Intriguingly, we find that vision-language pre-training on internet-scale datasets can tolerate \emph{extreme} batch sizes, \emph{e.g.} $B=1$M.
In Figure~\ref{fig:3}(a), we compare the loss behaviors when scaling the batch size for DP-Cap. We fix the effective noise $\sigma/B$ while varying the batch size.
In stark contrast to the previous observation from \cite{sander2023tan}, the loss trajectory is identical across different batch sizes. 
With this observation, we are able to successfully scale up the batch size for DP-Cap to as large as $B=1.3$M, achieving an effective noise of $5.6\times10^{-7}$, almost 10 times smaller than the effective noise of ViP in \citet{yu2023vip}.
Training DP-Cap under such a small effective noise allows it to extract information from the training dataset more efficiently under the DP constraint. 
In Figure~\ref{fig:3}, we show that with $B=1.3$M, the representation learned by DP-Cap even outperforms \emph{non-private} MAE trained on the same dataset. 
\revisioncolor{From these results, we find that the DP-Cap training can handle extreme large batches much more effectively than ViP in the context of differentially private training. This advantage is likely independent of image captioning models's inherent superiority over MAE. }

\begin{figure*}[t!]
    \begin{minipage}[c]{0.52\textwidth}
        \centering
        \includegraphics[width=\linewidth]{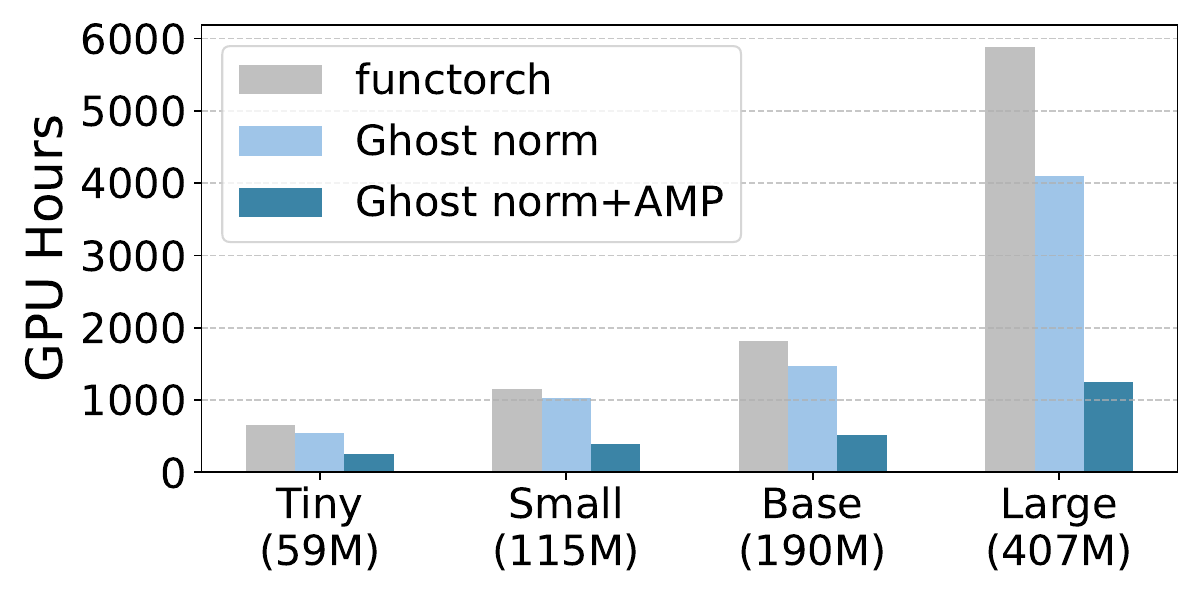}
        \vspace{-5ex}
        \caption{Number of GPU hours to train DP-Cap for a single epoch on 233M samples. For the Large model, we achieve a close to $5\times$ reduction.\label{fig:compute_cost}}
    \end{minipage}
    \hfill
    \begin{minipage}[c]{0.45\textwidth}
        \centering
        \resizebox{\linewidth}{!}{
        \begin{tabular}{c|cccc}
    	\toprule
    	\multirow{2}{*}{\textbf{Method}} & \multicolumn{4}{c}{\textbf{Model}} \\
            & Tiny & Small & Base & Large \\
            \midrule
    	\midrule
            \texttt{functorch} &  36 & 20 & 12 & 5 \\
            \cmidrule(lr){1-5} 
    	Ghost norm & 350 & 230 & 180 & 64 \\
            \cmidrule(lr){1-5} 
    	  Ghost norm+AMP & 420 & 310 & 225 & 90 \\
    	\bottomrule
        \end{tabular}
        }
        \captionof{table}{Max physical batch size under different per-sample gradient computation methods.\label{tab:max_batchsize}}
    \end{minipage}
    \vspace{-2ex}
\end{figure*}

\textbf{Improving training pipeline efficiency.} DP training is known to be computationally inefficient due to factors such as per-sample gradient computation~\citep{lee2020scaling, li2021large}. 
Training on internet-scale datasets using extreme batch sizes further complicates this issue. 
For example, a naive implementation of DP-Cap with per-sample gradient computation using \texttt{functorch} would take approximately 61 days(!) on 128 NVIDIA V100 GPUs. 
\revisioncolor{Recent studies~\citep{li2021large, bu2023differentially} have enhanced this efficacy by enabling DP training without the necessity to store per-sample gradients.}
We made significant efficiency improvements to the training pipeline using two techniques: ghost norm~\citep{li2021large} and the automatic mixed precision (AMP) package in PyTorch. 
Combining these two techniques with DP-SGD requires careful considerations to ensure both a correct DP guarantee as well as numerical stability. 
We detail the implementation in Appendix~\ref{app:computation}.

In Figure \ref{fig:compute_cost} we compare the compute cost of different gradient computation methods: \texttt{functorch}, ghost norm and ghost norm+AMP. The number of GPU hours is estimated on a single NVIDIA V100 GPU with 32GB memory using 100K samples. 
The improvement is especially notable for our largest model: Using ghost norm+AMP, we achieve a $4.7\times$ speedup compared to \texttt{functorch} and $3.3\times$ speedup compared to ghost norm alone, which amounts to a reduction from 61 days to 13 days when training for 32 epochs---a large but manageable compute cost. 
This improvement is due to both a more efficient forward-backward pass, as well as enabling a larger physical batch size; see Table \ref{tab:max_batchsize}. 
In addition, we adopt the TAN simulation framework~\citep{sander2023tan} to reduce the compute cost during hyperparameter search. 
Due to the batch size scaling behavior depicted in Figure \ref{fig:3}, TAN simulation is ideal for DP-Cap training and allows for rapid experimentation to identify promising methods before launching a full training run.

%% file: 4-experiments.tex
\section{Evaluation}\label{sec:evaluation}

We demonstrate the representation learning capabilities of DP-Cap on both vision ({\small\textsf{V}}) and vision-language ({\small\textsf{V-L}}) downstream tasks. For all evaluations, the DP-Cap model is first pre-trained using DP-SGD on a subset of LAION-2B~\citep{Schuhmann2022LAION5BAO}, and then fine-tuned non-privately on a downstream dataset.

\subsection{Downstream Tasks}\label{subsec:tasks}

\textbf{Linear probing ({\small\textsf{V}}).} We train a linear classifier on top of learned representations and evaluate its accuracy. We consider both full linear probing using the full downstream dataset, as well as few-shot linear probing, which subsamples the downstream dataset down to $K$ samples per class. Few-shot linear probing is especially useful for evaluating learned representations since the model must rely heavily on the generalizability of representations in order to perform well under data scarcity.

\textbf{Zero-shot image classification ({\small\textsf{V-L}})} is one of the most widely used methodologies for evaluating vision-language models~\citep{radford2021learning}. 
A strong zero-shot performance suggests that the image representation aligns well to text.
We perform zero-shot classification using the DP-Cap image encoder and text decoder by evaluating the likelihood of captions of the form ``this is a photo of a [label]''. We enumerate over different labels and predict the class that has the highest likelihood; see Section~\ref{sec:appendix_zero_shot} for full details.

\textbf{ARO (Attribution, Relation, and Order) ({\small\textsf{V-L}}).} The ARO benchmark~\citep{yuksekgonul2022and} can be used to gauge the adeptness of VLMs in understanding the compositional relationship between objects and attributes. A strong performance on ARO suggests that the learned image representation encodes semantic relationships such as ``the horse is eating the grass'' vs. ``the grass is eating the horse''.

\subsection{Experimental Setup}\label{subsec:exp-setup}
We present an overview of the experimental setup; refer to Appendix~\ref{sec:appendix-additional-details} for additional details.

\begin{table*}[t!]
	\centering
	\caption{Linear probing evaluation on downstream classification. Results for DP-NFNet, TAN, AlexNet and SimCLR are obtained from \citet{yu2023vip}. 
    For ViP \citep{yu2023vip}, we train with the same privacy parameters as for DP-Cap on the deduplicated dataset. More details are given in Appendix~\ref{sec:appendix_optim}.}
	\vspace{-0.05em}
	\resizebox{0.9\textwidth}{!}{
	    \footnotesize
		\begin{tabular}{llc|cccc}
			\toprule
			\multirow{1}{*}{Model} &   \multirow{1}{*}{pretraining data}  & \multirow{1}{*}{ DP? } 
			&   \multirow{1}{*}{ ImageNet-1K  } & \multirow{1}{*}{ Places-365 } & \multirow{1}{*}{ Places-205 } &  \multirow{1}{*}{iNat-2021}  
			  \\
            \midrule
		\midrule
		DP-NFNet  & ImageNet-1K &   {\large \textcolor{ao}{\ding{51}}}      &  45.3\%  &   40.1\% &  39.2\% &  28.2\% 
			\\ 
        \cmidrule(lr){1-7} 
		TAN   & ImageNet-1K & {\large \textcolor{ao}{\ding{51}}}    &   49.0\%  &  40.5\%  & 38.2\%  & 31.7\% 
			\\ 
        \cmidrule(lr){1-7} 
        {\color{gray} AlexNet }  & ImageNet-1K&{\large \textcolor{upsdellred}{\ding{55}}}  &   {\color{gray} 56.5\%}  &  {\color{gray} 39.8\%}  &  {\color{gray} 35.1\%} & {\color{gray} 23.7\%} 
			\\ 
        \cmidrule(lr){1-7} 
        {\color{gray} SimCLR }   & ImageNet-1K&{\large \textcolor{upsdellred}{\ding{55}}}  &  {\color{gray} 67.5\%}  & {\color{gray} 46.8\%}   & {\color{gray} 49.3\%}  & {\color{gray} 34.8\%}  
			\\ 
        \cmidrule(lr){1-7} 
		  {\color{gray} Cap} &  Dedup-LAION-233M &{\large \textcolor{upsdellred}{\ding{55}}} &  {\color{gray} 77.5\%} &  {\color{gray} 56.3\%} &  {\color{gray} 63.9\%}  &   {\color{gray} 63.9\%}
			\\ 
        \cmidrule(lr){1-7} 
		  {\color{gray} MAE} &  Dedup-LAION-233M &{\large \textcolor{upsdellred}{\ding{55}}} &  {\color{gray} 62.5\%} &  {\color{gray} 51.0\%} &  {\color{gray} 54.7\%}  &   {\color{gray} 42.3\%}
			\\ 
        \midrule
        \midrule
        {ViP}   & Dedup-LAION-233M &{\large \textcolor{ao}{\ding{51}}}   &  {56.5\%}  & {47.7\%}   & {49.6\%}  & {38.2\%}  
			\\ 
        \cmidrule(lr){1-7} 
        \cmidrule(lr){1-7} 

		  DP-Cap &  Dedup-LAION-233M & {\large \textcolor{ao}{\ding{51}}}  & {63.4\%} & {51.9\%}&  {54.3\%} & {44.5\%}
			\\ 
			\bottomrule
		\end{tabular}
	}
	\label{tab:main-table-linearprobing}
	\vspace{-0.2em}
\end{table*}

\begin{table*}[t!]
	\centering
	\caption{Performance of DP-Cap on zero-shot classification and compositional understanding (ARO). CLIP's zero-shot results are obtained from \citet{radford2021learning} (base model). For ARO, see Appendix~\ref{sec:appendix_ARO}.
 }
	\vspace{-0.1em}
	\resizebox{0.9\textwidth}{!}{
	    \footnotesize
		\begin{tabular}{lc|ccc|cccc}
			\toprule
			\multirow{2}{*}{Model}  & \multirow{2}{*}{ DP? } 
			& \multicolumn{3}{c|}{Zero-shot} 
			   & \multicolumn{4}{c}{ARO} 
			  \\[0.05em]
		  &  & ImageNet-1k  & CIFAR10 &  CIFAR100  & VGR  & VGA & COCO & Flickr 
			\\ 
            \midrule
        \midrule
    {Random Chance} & {-} & 0.1\%  &  10\% & 1\% & 50\% & 50\% & 20\% & 20\% 
			\\ 
        \midrule
    {\color{gray} CLIP } & {\large \textcolor{upsdellred}{\ding{55}}} &   {\color{gray}62.2\%} &  {\color{gray}91.3\%} &  {\color{gray}65.1\% }
    &  {\color{gray}62.4\%} &  {\color{gray}62.9\%} &  {\color{gray}47.8\%} &  {\color{gray}58.0\%}
			\\ 
    {\color{gray} Cap } & 
    {\large \textcolor{upsdellred}{\ding{55}}} &  {\color{gray}25.2\%}  &  {\color{gray}90.0\%}&  {\color{gray}37.4\%} &  {\color{gray}59.9\%} &  {\color{gray}87.2\%} &  {\color{gray}87.0\%} &  {\color{gray}87.4\%} 
			\\ 
    \midrule
    \midrule
    DP-Cap & {\large \textcolor{ao}{\ding{51}}} &  {7.8\%} & {54.4\%} & {16.4\%}  & {58.6\%} & {82.4\%} & {86.6\%}  & {87.2\%} 
			\\ 
			\bottomrule
		\end{tabular}
	}
	\label{tab:main-table-ARO}
	\vspace{-0.75em}
\end{table*}

\textbf{Datasets.} 
Following the approach introduced by \citet{yu2023vip}, we first pre-train on the Shader21k dataset~\citep{baradad2022procedural} of synthetic images.
We then train with DP-SGD on a subset comprising 233 million deduplicated (using SemDeDup~\citep{abbas2023semdedup}), NSFW-filtered and face-blurred (using an approach similar to \citet{yang2021obfuscation}) image-caption pairs from the (English-only) LAION-2B dataset~\citep{Schuhmann2022LAION5BAO}. We refer to this dataset as Dedup-LAION-233M.

We use the ImageNet-1K~\citep{Imagenet,Imagenet2}, CIFAR-10/100~\citep{krizhevsky2009learning}, Places-365/205~\citep{zhou2014learning} and iNaturalist-2021~\citep{van2021benchmarking} image classification datasets to assess the performance of learned image representations via full linear probing, few-shot linear probing, and zero-shot prediction. 
For vision-language tasks, we employ the Visual Genome Attribution (VGA), Visual Genome Relation (VGR), COCO-order~\citep{lin2015microsoft} and Flickr-30k~\citep{plummer2016flickr30k} datasets from the ARO benchmark~\citep{yuksekgonul2022and}. %
Finally, we evaluate image captioning using the MS-COCO 2017~\citep{lin2015microsoft} test set; result is shown in Figure~\ref{fig:1}(c) and Appendix~\ref{subsec:appendix-image-caption}.

\textbf{Model and training.} We use a transformer architecture~\citep{vaswani2017attention} for both the encoder and the decoder of DP-Cap, where the decoder applies causal cross-attention. 
\revisioncolor{We train DP-Cap model from scratch, including the embedding layer, with vocabulary size around 32,000.}
See Section~\ref{sec:appendix_optim} and \citet{tschannen2023image} for details. 
For privacy accounting we use R\'{e}nyi DP composition along with privacy amplification via Poisson subsampling~\citep{mironov2019r}, and convert to DP using \citet{balle2020hypothesis} through the Opacus library~\citep{yousefpour2021opacus}, targeting $\delta=1/N$ where $N$ represents the number of training samples. We refer to the non-private counterpart of DP-Cap trained on the same dataset as ``Cap''.

\subsection{Main Results}\label{subsec:eval-v-and-vl}

\begin{table*}[t!]
	\centering
	\caption{Ablation studies on the effect of dataset size and privacy budget $\varepsilon$ on DP-Cap (base). 
 }
	\vspace{-0.3em}
	\resizebox{\textwidth}{!}{
	    \footnotesize
		\begin{tabular}{llccc|cccc|cccc}
			\toprule
            \multirow{1}{*}{$\varepsilon$}
			& 
            \multirow{1}{*}{$\sigma$} & 
            \multirow{1}{*}{\# Data} &   
            \multirow{1}{*}{\# Steps} & 
            \multirow{1}{*}{$B$} 
			&   \multicolumn{4}{c|}{ImageNet-1K}   & \multicolumn{4}{c}{ARO ({\small\textsf{V-L}})}  
			  \\
            \midrule
		\midrule
		{ } & { } & { }   &  { } &  { }  & {0-shot} ({\small\textsf{V-L}})  & {1-shot} ({\small\textsf{V}}) &  {2-shot} ({\small\textsf{V}}) & {10-shot} ({\small\textsf{V}}) & {VGR} & {VGA} & {COCO} & {Flickr} 
			\\ 
  \midrule[\heavyrulewidth] 
		{\color{gray} $+\infty$} &
  {\color{gray} 0} &{\color{gray} 233M}   &  {\color{gray} 60,000}  &  {\color{gray} 40,960}     & {\color{gray}  {25.2\%} } &  {\color{gray} {27.0\%}} & {\color{gray} {37.2\%}} & {\color{gray} {57.9\%}} & {\color{gray} {59.9\%}} & {\color{gray} {87.2\%}} & {\color{gray} {87.0\%}}  & {\color{gray} {87.4\%}} 
			\\ 

        \cmidrule(lr){1-13} 
		 {8.0} & {0.728} &{233M} & {5708}  &  {1.3M}  &  
   \blue{7.8\%}    & \blue{10.3\%} &  \blue{15.6\%} & \blue{31.8\%} & \blue{58.6\%} & \blue{82.4\%} & \blue{86.6\%}
        & \blue{87.2\%} 
			\\ 

        \cmidrule(lr){1-13} 
		 {2.0} & {1.18} &{233M} & {2854} &  {1.3M}  &  \blue{3.2\% }    & \blue{7.0\%} &  \blue{10.8\%} & \blue{23.9\%} & \blue{58.5\% } & \blue{79.7\%} & \blue{85.3\% } & \blue{86.6\%} 
			\\ 

        \cmidrule(lr){1-13}  
		 {1.0} & {1.5} & {233M} &  {1427} &  {1.3M}  
   &  \blue{1.1\%}    & \blue{5.2\%} &  \blue{8.2\%} & \blue{19.9\%} & \blue{58.3\%} & \blue{75.6\%} & \blue{83.9\%} & \blue{85.3\%} 
			\\ 
        \cmidrule(lr){1-13} 
         8.0 & {0.728} & 23M  & {5708}&  {130K}   
        &  \blue{0.7\%}
        &  \blue{3.4\%}
        &  \blue{5.3\%}
        & \blue{13.3\%}  
        &  \blue{58.3\%}
        &  \blue{76.2\%}
        &  \blue{84.9\%}
        &  \blue{85.9\%} 
			\\ 
        \cmidrule(lr){1-13} 
         8.0 & {0.728} & 2.3M & 5708 &  {13K}  
        & \blue{0.1\%}    
        & \blue{1.8\%}
        & \blue{2.9\%}
        & \blue{8.1\%}
        & \blue{57.6\%}
        & \blue{66.4\%}
        & \blue{79.5\%}
        & \blue{82.0\%}
			\\ 
			\bottomrule
		\end{tabular}
	}
	\label{tab:eps-datasize}
	\vspace{-0.4em}
\end{table*}

\begin{table*}[t!]
	\centering
	\caption{Ablation studies on the effect of model size.
  We compare ViP and DP-Cap's number of encoder parameters. More details about the DP-Cap models can be found in Table~\ref{tab:appendix-table-cap-model-size}.
} %
	\vspace{-0.5em}
	\resizebox{\textwidth}{!}{
	    \footnotesize
		\begin{tabular}{llc|ccccc|cccc}
			\toprule
			\multirow{1}{*}{Model} & \multirow{1}{*}{Config} & 
            \multirow{1}{*}{\# parameters}  
			&  \multicolumn{5}{c|}{ImageNet-1K (Vision)}   & \multicolumn{4}{c}{ARO (Vision-Language)} 
			  \\
            \midrule
		\midrule
		{ } &   { }     & { }     & {1-shot} &  {2-shot} & {5-shot} & {10-shot} & \blue{LP} & {VGR} & {VGA} & {COCO} & {Flickr}
			\\ 
        \midrule[\heavyrulewidth] 
        {ViP} & Base &  86.6M  &  2.5\%    
        &  4.2\%
        & 8.5\%
        &  14.3\% 
        & \blue{56.5\%}
        & /  
        & /
        & /
        & /
			\\ 
        \midrule
		{DP-Cap}  & Tiny &  22.0M
        & \blue{7.9\%} 
        & \blue{12.1\%}
        & \blue{18.7\%}
        & \blue{25.2\%}
        & \blue{57.5\%}
        & \blue{58.6\%}
        & \blue{79.1\%}
        & \blue{85.7\%}
        & \blue{87.1\%}
			\\ 
        \cmidrule(lr){1-12} 
		{DP-Cap}  & Small &  49.0M
        & \blue{9.0\%} 
        & \blue{14.0\%}  
        & \blue{21.6\%}
        & \blue{28.9\%}
        & \blue{61.1\%}
        & \blue{59.1\%}
        & \blue{80.5\%}
        & \blue{86.0\%}
        & \blue{86.6\%}
			\\ 
        \cmidrule(lr){1-12} 
		{DP-Cap}  & Base &  86.6M
        & \blue{10.3\%}
        &  \blue{15.6\%}  
        & \blue{24.2\%}
        & \blue{31.8\%}  
        & \blue{63.4\%}
        &\blue{58.6\%}& \blue{82.4\%}& \blue{86.6\%}
        & \blue{87.2\%} 
			\\ 
        \cmidrule(lr){1-12} 
		{\blue{DP-Cap}}  & \blue{Large} &  \blue{407.3M}
        & \blue{11.8\%}
        &  \blue{17.5\%}  
        & \blue{26.2\%}
        & \blue{34.0\%}  
        & \blue{65.8\%}
        &\blue{59.5\%}& \blue{80.1\%}& \blue{86.6\%}
        & \blue{86.5\%} 
			\\ 
			\bottomrule
		\end{tabular}
	}
	\label{tab:scaling-model-size}
	\vspace{-0.4em}
\end{table*}

\textbf{Linear probing evaluation ({\small\textsf{V}}).} 
We assess the performance of the vision encoder on downstream tasks via linear probing.
In Fig~\ref{fig:1}(a), we compare the performance of DP-Cap and ViP \citep{yu2023vip} on ImageNet-1k few-shot linear probing. 
DP-Cap significantly improves over ViP, with up to $\times$2.5 better performance across different shots.
In addition, we evaluate the full linear probing accuracy of DP-Cap, ViP and other baselines in Table~\ref{tab:main-table-linearprobing}.
DP-Cap outperforms ViP and other DP models, including TAN~\citep{sander2023tan} and DP-NFNet~\citep{DeepMindUnlocking}, across all tasks.
DP-Cap even outperforms non-private AlexNet~\citep{krizhevsky2012imagenet} and except on ImageNet, SimCLR~\citep{simclr} (\blue{both were trained on ImageNet}).
We provide additional results for fine-tuning on downstream datasets in Table~\ref{tab:main-table-finetune-fewshot} (App.~\ref{sec:appendix-additional-exp-results}), also showing improvements over competing methods.

\textbf{Zero-shot performance ({\small\textsf{V-L}}).}
In the left three columns of Table~\ref{tab:main-table-ARO}, we evaluate the zero-shot performance of DP-Cap compared to non-private Cap and CLIP/BLIP on ImageNet-1k and CIFAR10/100.
Contrastive methods such as CLIP and BLIP have demonstrated greater suitability for zero-shot prediction compared to image captioning approaches~\citep{tschannen2023image}, which is evident by the disparity between the performance of Cap and CLIP/BLIP. 
Nevertheless, we observe that DP-Cap achieves noteworthy zero-shot classification performance that is significantly above random chance, and stands as the first DP model to do so. This accomplishment marks a promising milestone for DP training, although there remains a substantial performance gap between DP-Cap and Cap.

\textbf{Attribution, Relation, and Order (ARO) evaluation ({\small\textsf{V-L}}).} 
Contrastive-based methods such as CLIP often exhibit behavior akin to bag-of-words models~\citep{yuksekgonul2022and, tejankar2021fistful, basu2023augmenting}, making them less adept at performing well on the ARO benchmark. Remarkably, DP-Cap significantly outperforms \emph{non-private} CLIP in this context (see Fig~\ref{fig:1}(b) and Table~\ref{tab:main-table-ARO}), and even achieves performance close to that of \emph{non-private} Cap. Our result shows that DP training can be particularly effective for learning complex compositional relationships.

\vspace{-0.15cm}
\subsection{Ablation Studies}\label{subsec:dp-cap-scaling}

We perform ablation studies on the scaling behavior of DP-Cap with respect to the dataset size, privacy budget and model size. In Appendix~\ref{sec:appendix-additional-exp-results}, we show additional results on image captioning and on the impact of compute budget.

\textbf{Scaling dataset size.} We show that dataset scaling is crucial for effectively training DP-Cap as it results in better SNR under the same privacy budget (see Figure~\ref{fig:eps_N}).
We randomly subsample 1\% and 10\% of the Dedup-LAION-233M dataset, which is used for training our default DP-Cap-Base model in Table~\ref{tab:main-table-linearprobing} (denoted by Dedup-LAION-2M and Dedup-LAION-23M). 
We set the batch size to $B/100$ for Dedup-LAION-2M and $B/10$ for Dedup-LAION-23M, respectively. This allows the model to be trained for the same number of steps across the different datasets, although at a much larger effective noise level.
As shown in Table~\ref{tab:eps-datasize}, the number of training samples is critical for achieving strong performance for DP-Cap models: the zero-shot performance of our model trained on 1\% of the dataset achieves random zero-shot performance on ImageNet and much worse accuracy across the board on ARO.

\textbf{Impact of the privacy budget $\varepsilon$.}
We also investigate the performance of DP-Cap under lower privacy budgets ($\varepsilon=1$ and $\varepsilon=2$), employing the same batch size of 1.3 million. 
The outcomes of these experiments are presented in Table~\ref{tab:eps-datasize}. 
As anticipated, the utility of our model does exhibit a decline with decreasing $\varepsilon$. However, the performance degradation is relatively minor for the learned representation, with 10-shot ImageNet performance decreasing from 31.8\% ($\varepsilon=8$) to 19.9\% ($\varepsilon=1$).
More surprisingly, the performance impact on ARO is nearly negligible.
It is noteworthy that both models continue to outperform previous state-of-the-art DP models trained with $\varepsilon=8$ (see Figure~\ref{fig:1}). 
This phenomenon can be attributed to the relatively small effective noise resulting from the extreme batch size, which for $\varepsilon=1$ remains five times smaller than that used in \cite{yu2023vip}.

\textbf{Scaling model size.} 
Scaling up the model size is one of the most effective approaches for training better non-private foundation models~\citep{brown2020language, bommasani2021opportunities, touvron2023llama}. 
\revisioncolor{However, an intuitive understanding of DP-SGD~\footnote{This is because the added noise has $L_2$ norm $\approx \sigma C \sqrt{d} / B$, where $d$ is the number of model parameters, whereas the gradient norm is constrained to $C$ regardless of model size.} 
suggests that scaling up model size does not improve utility in DP training since more model parameters will lead to lower signal-to-noise ratio, especially when training large DP models from scratch}. 
\revisioncolor{Recent work~\citep{li2022does} demonstrates that during DP fine-tuning, gradients occupy a lower effective dimensional subspace, and larger networks can enhance the performance of DP-SGD.}
To test this hypothesis \revisioncolor{in pre-training}, we train DP-Cap with different model sizes (Tiny, Small, Base, Large) using the same hyperparameters and evaluate their performance in Table~\ref{tab:scaling-model-size},\ref{tab:zero-shot-model-size-table-appendix}; see Table~\ref{tab:appendix-table-cap-model-size} for details about different model sizes.
We observe consistent improvements when scaling up the model from {DP-Cap-Tiny} to {DP-Cap-Large}.
Our observation suggests that DP-Cap has strong model scaling behavior even with DP-SGD training.

%% file: 5-discussion.tex
\vspace{-0.2cm}
\section{Discussion and Future Work}\label{sec:discussion}

We demonstrated that DP representation learning via image captioning is viable. In particular, image captioning is an ideal objective that supports both per-sample loss and large batch training---two critical ingredients in DP-SGD. When applied to the Dedup-LAION-233M dataset, the trained model learns useful image representations for downstream tasks and exhibits strong multi-modal capabilities. 

Through our study we also identify three open problems in the general direction of DP pre-training of large-scale foundation models that are difficult to handle with existing techniques:

\begin{enumerate}[leftmargin=*]
\item Scaling up the batch size to extreme levels is crucial for reducing the effective noise and facilitating model convergence. Is there a fundamental reason why image captioning can support extremely large batch training?
When the model does not scale well with batch size, is it due to poor optimization or is it inherent to the model and/or task? Answers to these question can help identify other successful training recipes that further improve the quality of DP-learned representations.
\item
Scaling up the model size is typically beneficial in non-private representation learning, but can be ineffective in DP training due to a decrease in gradient SNR. 
While we have observed a performance improvement from Tiny to Large, we anticipate that deterioration could occur when scaling up to a much larger model size, especially when the signal-to-noise ratio (SNR) undergoes a certain threshold.
Does this threshold exist? Can this trade-off be resolved through parameter-efficient architectures?
\item
Contrastive learning offers unique advantages compared to other multi-modal learning objectives such as learning image and text representations that are semantically aligned, but is not compatible with standard DP-SGD training. What techniques can enable differentially private contrastive learning?
\end{enumerate}

%% file: sec_appendix.tex
\section{Implementation Details}\label{sec:appendix-additional-details}

\subsection{Training Details}\label{sec:appendix_optim}

\textbf{DP accounting.} 
We use RDP accounting with subsampling from the Opacus library~\citep{yousefpour2021opacus}. Let $D_{\alpha}$ denote the R\'{e}nyi divergence of order $\alpha$~\citep{renyi1961measures}, and let
\begin{equation}
   g_\alpha(\sigma,q) := D_{\alpha}{((1-q)\gauss(0,\sigma^2)+q\,\gauss(1,\sigma^2)  \,\|\,  \gauss(0,\sigma^2))}.
\end{equation}
Then, from \citet{mironov2019r}, performing $S$ steps of DP-SGD satisfies $(\varepsilon, \delta)$-DP with:
\begin{align}
\label{eq:eps_rdp}
\varepsilon := \min_\alpha \Big\{ S\cdot g_\alpha(\sigma,q) + \frac{\log(1/\delta)}{\alpha-1}\Big\}.
\end{align}
The quantity $g_\alpha(\sigma,q)$ can be upper bounded mathematically or derived numerically: we use the Opacus~\citep{yousefpour2021opacus} library for accounting in our work.

{Regarding the DP guarantee, $\varepsilon$-DP bounds the amount of information extracted from each training sample by $\varepsilon$. Notably, for DP-Cap, each sample is made of \{image + caption\}, while ViP utilizes \{image\} only. Consequently, DP-Cap inherently offers an equivalent or better privacy guarantee for each image. One way to see it is to note that DP provides protection against membership inference attacks~\citep{shokri2017membership}. 
Suppose $\varepsilon$-DP upper bounds the success rate of a membership inference attack (when given the image-text pair) against DP-Cap as $\leq p$. Then the MIA success rate when given only the image can be at most $p$ since the attacker has strictly less information. 
This is exactly the upper bound for the success rate of a membership inference attack against ViP. In other words, any attacker that can attack the image+caption model (such as DP-Cap) can also attack the image-only model (such as ViP).} 

{
On the other hand, since the \{image+caption\} models utilize the caption, the privacy leakage from the text part of the image-caption pair is non-zero for $\varepsilon>0$. 
It is worth noting that in our set up since we use DP, we protect the captions with the same $\varepsilon$-DP guarantee. 
Thus, the privacy protection for DP-Cap is neither strictly stronger nor strictly weaker than that for ViP, so the two privacy notions are not directly comparable.
}

\textbf{Model details and task description.} We utilize a transformer architecture \citep{vaswani2017attention} DP-Cap.
This captioning model uses a text decoder that generates captions in an auto-regressive manner, utilizing a full attention mechanism on the vision encoder's output, as well as causal attention on the text.
This architecture is closely aligned with the Cap architecture introduced in \cite{tschannen2023image}. See Table \ref{tab:appendix-table-cap-model-size} for details about the transformer architecture for different sizes. All results utilize the \emph{base} model with the exception of the comparison in Table~\ref{tab:appendix-table-cap-model-size}.

\textbf{Hyperparameters.}  Our choice of gradient clipping factor is $C=1$, as we did not observe any performance improvement with other values. We always use AdamW \citep{loshchilov2018decoupled} for training. 
We use a learning rate of $5.12\times10^{-4}$.
The learning rate is kept constant across batch sizes for TAN simulations and for the performance comparison in Figure~\ref{fig:3} as the effective noise is kept constant in these cases \citep{sander2023tan}.
We use a maximum length of $40$ tokens to process the LAION captions.
We use a linear schedule, with $40\%$ of warm-up iterations, and $2\times$ the entire training as decay horizon.
As opposed to what was previously observed \citep{DeepMindUnlocking, sander2023tan}, the learning rate schedule played an important role for us with DP-SGD training.
We use a weight decay of $0.05$.
These choices come from hyperparameter search using TAN simulation with our base model.
Following the standard practice \citep{berrada2023unlocking, DeepMindUnlocking, li2021large, yu2023vip, sander2023tan}, we do not count hyperparameter search within our privacy budget. \cite{liu2019private} have shown that hyperparameter search might not incur observable privacy loss.

\textbf{Pre-training DP-Cap on the synthetic dataset.} 
Compared to the encoder and decoder architecture design used in masked autoencoders (MAE)~\citep{he2022masked}, the two main differences of the image captioning model used in this paper are:  
(1) The output of the encoder is fed into the decoder via cross-attention~\citep{vaswani2017attention} in Cap;
and (2) The self-attention used in the Cap decoder is causal self-attention.
Similar to \citet{yu2023vip}, we apply the synthetic image dataset, Shaders21K~\citep{baradad2022procedural}, to pre-train the DP-Cap model via MAE-based training.
We follow most of the training setups used in ViP synthetic pre-training~\citep{yu2023vip}, except that we feed the output of the encoder to the decoder via cross-attention.
The training loss of the synthetic pre-training in this paper is still the reconstruction loss used in MAE~\citep{he2022masked}, and we did not leverage real-world text data for pre-training. 
After the model is pre-trained on Shaders21K, we change the self-attention to causal self-attention in the decoder, and replace the final layer (for pixel-wise reconstruction) of the decoder with the (randomly initialized) decoding layer for next word prediction.
After making these modifications, we apply DP-SGD to pre-train our DP-Cap model with standard image captioning training objectives (see Section~\ref{subsec:pretrain-captioners-with-DPSGD}).

\begin{table*}[h]
	\centering
	\caption{ Details of transformer backbone variants used in DP-Cap.
 }
	\vspace{-0.05em}
	\resizebox{0.98\textwidth}{!}{
	    \footnotesize
		\begin{tabular}{l|cccc|c}
			\toprule
			\multirow{1}{*}{Model} & \multirow{1}{*}{Encoder depth} &   \multirow{1}{*}{Encoder width} 
			& \multirow{1}{*}{Decoder depth} & \multirow{1}{*}{Decoder width} 
			&   \multirow{1}{*}{\# parameters (encoder \& decoder)}
			  \\
            \midrule
		\midrule
        {DP-Cap-Tiny} &  12  & 384  & 6 & 384  &  59M
			\\ 
        \cmidrule(lr){1-6} 
	{DP-Cap-Small} & 12 & 576  & 6  & 576 & 115M
			\\
        \cmidrule(lr){1-6} 
	  {DP-Cap-Base} & 	  12  & 768  & 6  & 768 &  190M
			\\ 
        \cmidrule(lr){1-6} 
	  {DP-Cap-Large} & 	  24  & 1024  & 6  & 768 &  407M
			\\ 
			\bottomrule
		\end{tabular}
	}
	\label{tab:appendix-table-cap-model-size}
\end{table*}

\textbf{Pre-training ViP.} To conduct a comparison with training on an identical datasets, we follow the methodology outlined in \citep{yu2023vip} to train with DP-SGD a MAE-based model, but with a change in the training data from LAION-233M to Dedup-LAION-223M, and use the same encoder's synthetic initialization as for DP-Cap. 
We further examine the linear probing performance on ImageNet and observe a 2\% between the original model and the one trained on the deduplicated dataset. 
In addition, to corroborate the observation made in Figure~\ref{fig:3}, which suggests that the MAE-based method struggles to effectively harness massive batch sizes for achieving low effective noise in DP-SGD, we also train ViP models with larger batches, up to using the exact privacy parameters employed for DP-Cap (under $\varepsilon=8$) with a notably large batch size of 1.3 million, and showcase the results in Table~\ref{tab:appendix_vip}.
\blue{For full linear probing, we observe only a small improvement over the original ViP model that was trained with batch size 98k}. The success of DP-Cap is not solely attributed to its appropriate privacy parameters but is also a consequence of its remarkable ability to \blue{leverage the small effective noised induced by} extremely large batch sizes.

\subsection{Computation cost}\label{app:computation}

\paragraph{Mixed Precision Package \& Ghost Norm}
DP-SGD introduces additional computational overhead compared to non-private training, primarily due to the computation of per-sample gradient norms. By employing the ghost norm technique of~\citet{li2021large}, we have successfully reduced the computational cost by up to one third with the Large Model (see Figure~\ref{fig:compute_cost}) compared to using \texttt{functorch}.
The \texttt{torch.amp} package offers convenient methods for mixed precision, significantly speeding up operations like linear layers. However, it often leads to NaNs due to low precision handling of extreme values. While one can skip a step that led to NaNs in normal training, combining AMP with Ghost Norm is more complex.
Ghost Norm requires two backward passes. In the first pass, per-sample gradient norms are computed. If one gradient norm is NaN, it contaminates the entire batch, leading to a NaN in the second backward pass. This issue is particularly prevalent in our setting with a batch size of $1.3$M, as even a minuscule proportion of computations leading to NaNs can cause problems. To address this, we propose two solutions:

\begin{itemize}
    \item \textbf{Loss Scaler}: We employ a similar trick to the standard use of AMP to reduce the number of NaNs. This involves dynamically upscaling and downscaling the loss with \texttt{torch.cuda.amp.GradScaler}. The same factor is used before the first and the second backward, and is updated based on the outputs of the second backward only.
    \item \textbf{Clipping to 0}: If any per-sample gradient norm computation results in a NaN value after the first backward, we set its clipping coefficient (the multiplicative coefficient in front of the corresponding per-sample loss for the second backward, as detailed in~\citet{li2021large}) to 0 for the second backward. In this case, we do not update the loss scaling factor.
\end{itemize}

It's worth noting that the second solution is entirely valid for DP: instead of clipping the per-sample gradient to a norm $C$, it clips it to 0 in cases where computation results in a NaN value. 
This approach effectively mitigates the issue of NaN contamination in large batches. 
Overall, We have successfully reduced the computational cost by a factor 5 for the Large Model compared to \texttt{functorch}.

\paragraph{TAN simulation} Crucially, to achieve a favorable privacy-utility trade-off, DP-SGD necessitates training with massive batches over a substantial number of steps to achieve a good privacy-utility trade-off, as elaborated in Section~\ref{subsec:effective_strategies_DPCAP}.
All our hyperparameter search were performed using the TAN simulation~\citep{sander2023tan} for one epoch on our Dedup-LAION-233M.
For our $\varepsilon=8$ models, we limited training to 32 epochs, a process that took 5 days utilizing 128 V100 GPUs for the Base model.

While we have tried to reduced it as much as possible, training DP-Cap imposed a considerable energy consumption, resulting in elevated CO2 emissions.
Our intention in releasing these models is to contribute to the mitigation of future carbon emissions, as the training has already been completed.

\begin{table*}[t!]
	\centering
	\caption{Set-ups for our training for ViP, MAE \citep{he2022masked} and DP-Cap: ImageNet-1k linear probing.}
	\vspace{-0.25em}
	\resizebox{0.9\textwidth}{!}{
	    \footnotesize
		\begin{tabular}{lcc|ccccc}
			\toprule
			\multirow{2}{*}{Model} &\multirow{2}{*}{pretraining data} & \multirow{2}{*}{($B$, $\sigma$, $S$)} & 
            \multicolumn{5}{c}{ImageNet-1K}   
			  \\ [0.05em]
     & &  & 1-shot & 2-shot & 5-shot & 10-shot & full
      \\
        \midrule[\heavyrulewidth] 
        ViP \citep{yu2023vip} &LAION-233M & (98k, 0.48, 6000)  
        &  2.5\%
        &  4.1\%
        &  8.5\%
        &  14.2\%
        &  55.7\%
			\\ 
        \cmidrule(lr){1-8} 
	ViP &	Dedup-LAION-233M &  (98k, 0.474, 5708)
        &  2.3\%    
        &  3.9\%
        & 8.0\%
        &  13.6\% 
        & 53.4\%
			\\ 
        \cmidrule(lr){1-8} 
	ViP &	Dedup-LAION-233M &  (200k, 0.513, 5708)
        & 2.5\%    
        &  4.1\%
        & 8.3\%
        &  14.1\% 
        & 54.0\%
			\\ 
        \cmidrule(lr){1-8} 
	ViP &	Dedup-LAION-233M &  (400k, 0.564, 5708)
        & 2.5\%    
        &  4.2\%
        & 8.7\%
        &  14.7\% 
        & 55.2\%
			\\ 
        \cmidrule(lr){1-8} 
	ViP &	Dedup-LAION-233M &  (1.3M, 0.728, 5708)
        &  \blue{2.7\%}
        &  \blue{4.6\%}
        &  \blue{9.4\%}
        &  \blue{15.7\%}
        & \blue{56.5\%}
        \\ 
           \cmidrule(lr){1-8} 
		\color{gray}MAE (Non private) &   \color{gray}Dedup-LAION-233M &  \color{gray}(40960, 0, 40000)
        &  \color{gray}3.4\%
        &  \color{gray}5.8\%
        &  \color{gray}11.8\%
        &  \color{gray}19.5\%
        & \color{gray}62.5\%
			\\ 
           \cmidrule(lr){1-8} 
		DP-Cap & Dedup-LAION-233M &  (98k, 0.474, 5708)
        &  \blue{4.2\%}
        &  \blue{6.9\%}
        &  \blue{11.4\%}
        &  \blue{17.2\%}
        & \blue{50.6\%}
			\\ 
           \cmidrule(lr){1-8} 
		DP-Cap &   Dedup-LAION-233M &  (200k, 0.513, 5708)
        &  \blue{5.6\%}
        &  \blue{8.9\%}
        &  \blue{14.5\%}
        &  \blue{20.2\%}
        & \blue{54.2\%}
			\\ 
           \cmidrule(lr){1-8} 
		DP-Cap &   Dedup-LAION-233M &  (400k, 0.564, 5708)
        &  \blue{7.6\%}
        &  \blue{11.5\%}
        &  \blue{18.5\%}
        &  \blue{25.3\%}
        & \blue{59.1\%}
            \\
           \cmidrule(lr){1-8} 
		DP-Cap &   Dedup-LAION-233M &  (1.3M, 0.728, 5708)
        &  \blue{10.3\%}
        &  \blue{15.6\%}
        &  \blue{24.2\%}
        &  \blue{31.8\%}
        & \blue{63.4\%}
			\\ 

			\bottomrule
		\end{tabular}
	}
	\label{tab:appendix_vip}
\end{table*}

\subsection{Evaluation Details}

\subsubsection{Details about Zero-shot Image Classification}\label{sec:appendix_zero_shot}

While methods employing contrastive learning, such as CLIP, excel in this task, captioning methods exhibit comparatively lower performance, and with greater computational demands during evaluation.
To evaluate a captioning model's zero-shot performance, we employ two distinct strategies:

\begin{itemize}[leftmargin=*]
    \item 
    \textbf{Tree-based search}: We initiate caption generation with a prompt like ``this is a photo of a," and greedily select the most likely next token among those that lead to valid completions within the true label set.
    The process continues until an End of Sentence (EOS) token is reached. 
    For instance, if there are only two labels starting with ``car": ``car [EOS]" and ``carpet [EOS]", and the initial predicted token is ``car". 
    Then the text decoder will predict the next token among ``[EOS]" and ``pet".
    If, among these two, ``[EOS]" is chosen, and ``car [EOS]" corresponds to the true label, then the zero-shot prediction is deemed correct.
    \item \textbf{Loss-based classification}: We assess, for each image, the probability of various captions that begin with ``this is a photo of a [...]" where ``[...]" is substituted with all feasible labels. Subsequently, we select the label that yields the most probable caption.
\end{itemize}

The ``loss-based classification" comes with significantly higher computation costs as all the different captions have to be evaluated for each image (there representations is conditional to the image). 
For ImageNet, it implies 1000 forwards through the decoder for each image. 
We thus employ the tree-based search for presenting our findings in Table~\ref{tab:main-table-ARO}, although its greedy character with no backtracking is not optimal.
Surprisingly, our preliminary experiments suggest the tree-based search gives comparable results.

\subsubsection{Details about ARO Evaluation}\label{sec:appendix_ARO}
We adhered to the protocol and code base established in \cite{yuksekgonul2022and} for re-evaluating CLIP's performance, and we observe slightly different results (see Table~\ref{tab:appendix-table-ARO}).
For our captioning models, our approach involved computing the cross-entropy loss for all possible captions associated with each image and subsequently selecting the one with the lowest loss.

\begin{table*}[t!]
	\centering
	\caption{Compositional understanding (ARO): Results for CLIP (base) in  \citet{yuksekgonul2022and} compared to our evaluation.
 }
	\vspace{-0.1em}
	\resizebox{0.7\textwidth}{!}{
	    \footnotesize
		\begin{tabular}{l|cccc}
			\toprule
			\multirow{2}{*}{Model} 
			   & \multicolumn{4}{c}{ARO} 
			  \\[0.05em]
		 & VGR  & VGA & COCO & Flickr 
			\\ 
            \midrule
        \midrule
    {CLIP (eval from \citet{yuksekgonul2022and})}  &59\% &63\% &46\% & 60\%
			\\ 
    {CLIP (our eval)} & 
    62.4\% &62.9\% &47.8\% & 58.0\%
			\\ 
			\bottomrule
		\end{tabular}
	}
	\label{tab:appendix-table-ARO}
	\vspace{-0.5em}
\end{table*}

\subsubsection{Details about Linear Probing and Fine-tuning Evaluation.}
Few-shot linear probing is accomplished using the Cyanure library~\citep{mairal2019cyanure}. We use the same hyper parameters as in \citet{assran2022masked}. We adapted the MAE \citep{he2022masked} code base for full linear probing, and we use the same hyperparameters as in \cite{yu2023vip} (extract 12 layers of the image encoder, LARS optimizer \citep{LARS} with base learning rate of 0.1, no weight decay and batch size of 16384).

\section{Additional Results}\label{sec:appendix-additional-exp-results}

\subsection{Additional experiments}

{\textbf{Impact of the initialization (V).}
Our synthetic initialization for DP-Cap achieves less favorable results than the one from ViP reaches 50\% \citep{yu2023vip}; for instance, for full linear probing on ImageNet, they achieve 44\% (Figure~\ref{fig:benefit-syn-pretrain}) and 50\% respectively.
However we have demonstrated that training with DP on top of synthetic initialization leads to significantly better results for DP-Cap compared to ViP for all the metrics; see Table~\ref{tab:main-table-linearprobing}, Table~\ref{tab:main-table-finetune-fewshot} and Figure~\ref{fig:1}. 
We observe that this superiority also appears when the models are trained from random initialization: as shown in Table~\ref{tab:appendix_randominit}, the improvement over ViP is even larger when training without synthetic initialization.}

\begin{table*}[t!]
	\centering
	\caption{Training from random initialization: Superiority of DP-Cap over ViP, both trained from random initialization.}
	\vspace{-0.25em}
	\resizebox{0.5\textwidth}{!}{
	    \footnotesize
		\begin{tabular}{c|cccc}
			\toprule
			 \multirow{2}{*}{Model} & 
            \multicolumn{4}{c}{ImageNet-1K}   
			  \\ [0.05em]
      &  1-shot & 2-shot & 10-shot & full
      \\
        \midrule[\heavyrulewidth] 
         ViP ($\varepsilon=8$) 
        &  \blue{0.1\%}
        &  \blue{1.7\%}
        &  \blue{6.1\%}
        &  \blue{23.9\%}
			\\ 
		  DP-Cap ($\varepsilon=8$)
        &  \blue{5.6\%}    
        &  \blue{8.5\%}
        &  \blue{18.8\%} 
        & \blue{47.0\%}
			\\ 
			\bottomrule
		\end{tabular}
	}
	\label{tab:appendix_randominit}
\end{table*}

{\textbf{Fine-tuning (V).} In Table~\ref{tab:main-table-finetune-fewshot}, we present DP-Cap's performance in fine-tuning for few-shot evaluation. In contrast to the linear probing results shown in Table~\ref{tab:main-table-linearprobing}, the network is completely unfrozen. Therefore, we assess DP-Cap's capabilities primarily as a network initialization. Similarly to the linear probing results, we note a significant improvement in all metrics compared to previous DP vision backbones. Note that, similarly to linear probing comparison in Figure~\ref{fig:1}, we compare to non-private model performance which provides information about the performance gap between private models and non-private models.
For fair comparison, we evaluate on the same same datasets than~\citet{yu2023vip}.

\begin{table*}[t!]
	\centering
	\caption{Fine-tuning evaluation on few-shot downstream classification.
 }
	\vspace{-0.1em}
	\resizebox{0.9\textwidth}{!}{
	    \footnotesize
		\begin{tabular}{c|ccc|ccc|ccc}
			\toprule
			\multirow{2}{*}{Model} 
			& \multicolumn{3}{c|}{Aircraft} 
			   & \multicolumn{3}{c|}{ Caltech-101 } & \multicolumn{3}{c}{ CIFAR-100 } 
			  \\[0.05em]
		    & 10-shot  & 20-shot &  30-shot  & 5-shot & 10-shot  & 30-shot  & 5-shot & 10-shot  & 30-shot 
			\\ 
            \midrule
        \midrule
    {\color{gray} AlexNet } &    {\color{gray} 23.3\% } & 
    {\color{gray} 34.4\% } & {\color{gray} 41.4\% } 
    & {\color{gray} 64.7\% } & 
    {\color{gray} 73.6\% } & {\color{gray} 81.4\% } 
    & {\color{gray} 29.7\% } & 
    {\color{gray} 36.3\% } &{\color{gray} 49.3\% } 
			\\ 
        \cmidrule(lr){1-10} 
    {\color{gray} SimCLR } &    {\color{gray} 38.8\% } & 
    {\color{gray} 56.9\% } & {\color{gray} 64.9\% } 
    & {\color{gray} 81.7\% } & 
    {\color{gray} 89.1\% } & {\color{gray} 94.5\% } 
    & {\color{gray} 49.9\% } & 
    {\color{gray} 60.2\% } &{\color{gray} 71.8\% } 
			\\ 
        \cmidrule(lr){1-10} 
    {TAN} &    22.8\% & 37.9\% &  46.0\% &  49.3\% &  66.4\% & 77.9\% & 21.3\% & 27.8\% & 42.4\% \\
    \midrule
    \midrule
    {ViP} &    31.6\% & 53.1\% & 64.3\% &68.1\% &79.0\% &88.9\% &30.7\% &41.0\% &57.5\% \\
    \midrule
    DP-Cap &   \blue{37.5\%}   & \blue{57.9\%}   &  \blue{66.7\%} &  \blue{70.3\%} & \blue{81.3\%}  & \blue{90.0\%}  &  \blue{36.3\%}&  \blue{46.3\%} &  \blue{62.1\%}
			\\ 
			\bottomrule
		\end{tabular}
	}
	\label{tab:main-table-finetune-fewshot}
	\vspace{-0.4em}
\end{table*}

\begin{table*}[t!]
	\centering
	\caption{Ablation studies on the effect of model size for zero-shot prediction.
 }
	\vspace{-0.1em}
	\resizebox{0.85\textwidth}{!}{
	    \footnotesize
		\begin{tabular}{llcc|ccc}
			\toprule
			\multirow{2}{*}{Model}  & \multirow{2}{*}{Config} & \multirow{2}{*}{\# parameters} &  \multirow{2}{*}{ DP? } 
			& \multicolumn{3}{c}{Zero-shot} 
			  \\[0.05em]
		  &  &  &  & ImageNet-1k  & CIFAR10 &  CIFAR100   
			\\ 
            \midrule
        \midrule
    {Random Chance} & {-} & {-}  & {-} & 0.1\%  &  10\% & 1\% 
			\\ 
        \midrule
    {\color{gray} Cap } &  Base & 86.6M &
    {\large \textcolor{upsdellred}{\ding{55}}} & {\color{gray} 25.2\%}  & {\color{gray} 90.0\%}& {\color{gray} 37.4\%} 
			\\ 
    \midrule
    \midrule
    DP-Cap & Tiny & 22.0M & {\large \textcolor{ao}{\ding{51}}} &  {5.0\%} & {46.5\%} & {11.1\%}  
			\\ 
    DP-Cap & Small & 49.0M & {\large \textcolor{ao}{\ding{51}}} &  {6.9\%} & {53.6\%} & {17.1\%}  
			\\ 
    DP-Cap & Base &  86.6M & {\large \textcolor{ao}{\ding{51}}} &  {7.8\%} & {54.4\%} & {16.4\%}  
			\\ 
    DP-Cap & Large & 407.3M  & {\large \textcolor{ao}{\ding{51}}} &  {9.2\%} & {62.1\%} & {24.0\%}  
			\\ 
			\bottomrule
		\end{tabular}
	}
	\label{tab:zero-shot-model-size-table-appendix}
	\vspace{-0.25em}
\end{table*}

\textbf{Captioning task ({\small\textsf{V-L}}).} 
We evaluate the image captioning performance of DP-Cap in comparison to non-private Cap.
In Fig.~\ref{fig:1}(c), we present some (randomly chosen) captions generated by DP-Cap; more examples for DP-Cap and Cap can be found in Appendix~\ref{subsec:appendix-image-caption}. 
Qualitatively, DP-Cap seems to generate reasonably good captions, similar to the ones generated by Cap. 
We also compare the two models quantitatively using the CIDEr metric~\citep{vedantam2015cider} to evaluate the generated captions on the MS-COCO test set, and the results are summarized in the last column of Table~\ref{tab:main-table-ARO}. 
As DP-Cap and Cap are only trained on noisy captions from LAION, the CIDEr metric on MS-COCO is relatively low for both models. Moreover, despite the similar performance between DP-Cap and Cap on ARO, the gap is much more significant for the captioning evaluation. Given these results, it is plausible that even though DP-Cap attains remarkably compositional understanding capabilities, its ability to generate text is still limited.

{We also fine-tune Cap and DP-Cap's decoders (while freezing the encoder) for one epoch on the MS-COCO train set, and assess the improvement in CIDEr scores in Table~\ref{tab:appendix_captioning_eval} to showcase the quality of the image representations and decoder initialization from the pre-training stage. 
The captions in Figure~\ref{fig:1} and Appendix~\ref{subsec:appendix-image-caption} are generated using models that were \emph{not} trained on MS-COCO.

\begin{table*}[t!]
	\centering
	\caption{Captioning evaluation on the MS-COCO test set of Cap and DP-Cap. For ``fine-tuned", the model's decoder is fine-tuned for one epoch on the MS-COCO train set (with the image encoder frozen).}
	\vspace{-0.25em}
	\resizebox{0.3\textwidth}{!}{
	    \footnotesize
		\begin{tabular}{l|cc}
			\toprule
			\multirow{2}{*}{Model} &
            \multicolumn{2}{c}{CIDEr score}   
			  \\
        {} & original & fine-tuned
        \\
        \midrule[\heavyrulewidth] 
        Cap & 29.9   
        &  79.2
			\\ 
        \cmidrule(lr){1-3} 
		DP-Cap &  15.7
        & 51.3
			\\ 
			\bottomrule
		\end{tabular}
	}
	\label{tab:appendix_captioning_eval}
\end{table*}

\textbf{What can we do with more compute budget?} We restricted training the DP-Cap model for a compute budget of 32 epochs on the Dedup-LAION-233M dataset for each of our models with $\varepsilon=8$.
To fit the privacy budget while utilizing a batch size of 1.3 million and training for 32 epochs, RDP analysis yields $\sigma=0.728$. 
However, we anticipate that further increasing the compute budget can yield even better models up to a certain limit:
With the same $\varepsilon$ and batch size, doubling the compute to 64 epochs only necessitates a 12\% increase in $\sigma$. 
This increase enables twice as many steps to be performed with only a marginal increase in effective noise, potentially allowing the model to converge to a better solution.

In the absence of necessary compute for running this experiment, \emph{we partially validate this hypothesis through the Total Amount of Noise (TAN) simulation}, training for the same number of gradient steps and with the same SNR per step, but using a $\times$32 smaller batch size and $\times$32 smaller $\sigma$ to simulate at $\times$32 lower compute.
Our results in Table~\ref{tab:TAN_compute_simulation} indicate a significant performance improvement of 5\% in 10-shot accuracy on ImageNet (compared to a similar simulation of the 32 epochs training). 
However, increasing the budget further to 128 epochs does not seem to enhance performance compared to the 64 epoch counterpart. 
Intuitively, the lower gradient SNR and larger number of gradient steps have opposite effects on optimization, and pushing past the ``sweet spot'' of training for 64 epochs at $\sigma=0.81$ results in noisy steps that are unproductive for model convergence. 
To surpass the performance of the 64-epoch, 1.3-million batch size DP-Cap model, training with an even larger batch size appears necessary. We emphasize again that this result is derived through TAN simulation, and actual, compute-intensive training is required to fully validate this assertion.

\begin{table*}[t!]
	\centering
	\caption{TAN simulation of the impact of the compute budget on the performance at fixed $B$.}
	\resizebox{0.45\textwidth}{!}{
	    \footnotesize
		\begin{tabular}{c|ccc}
			\toprule
			\multirow{2}{*}{}  &  \multicolumn{2}{c}{$\sigma$}  \\[0.05em]
		    &  0.81 &  0.95 
			\\ 
            \midrule
        \midrule
    {Epochs} &  64 ($\times$2) & 128 ($\times$4)
			\\ 
    {Effective noise $\sigma/B$} & $\times$1.12 & $\times$1.32
            \\
    {Predicted Final loss} & $-$0.2 & $-$0.2 
            \\
    {Predicted 10-shot ImageNet} & $+$5\% & $+$5\% 
			\\ 
			\bottomrule
		\end{tabular}
	}
	\label{tab:TAN_compute_simulation}
\end{table*}

\subsection{More on the Impact of dataset size and privacy parameters}
\label{subsec:appendix-dataset-scaling}

\textbf{Dataset size.} We emphasize here (again) the importance of having enough training data to achieve a good privacy-utility trade-off with DP-SGD.
As depicted in Figure~\ref{fig:eps_N}, increasing the number of training samples $N$ while keeping the same number of equivalent DP-SGD steps (\emph{i.e.}, keeping batch size $B$, noise $\sigma$, and number of update steps $S$ constant) considerably reduces the privacy budget $\varepsilon$.
Equivalently, having more data allows for an increase in the number of equivalent DP-SGD steps at fixed $\varepsilon$.
Similar observations were also made by \citet{tramer2020differentially,mcmahan2017learning}.
The abundance of pre-training data available for training foundation models thus proves highly compatible with DP requirements.

\begin{figure}[t!]
     \centering
\includegraphics[width=0.8\textwidth]{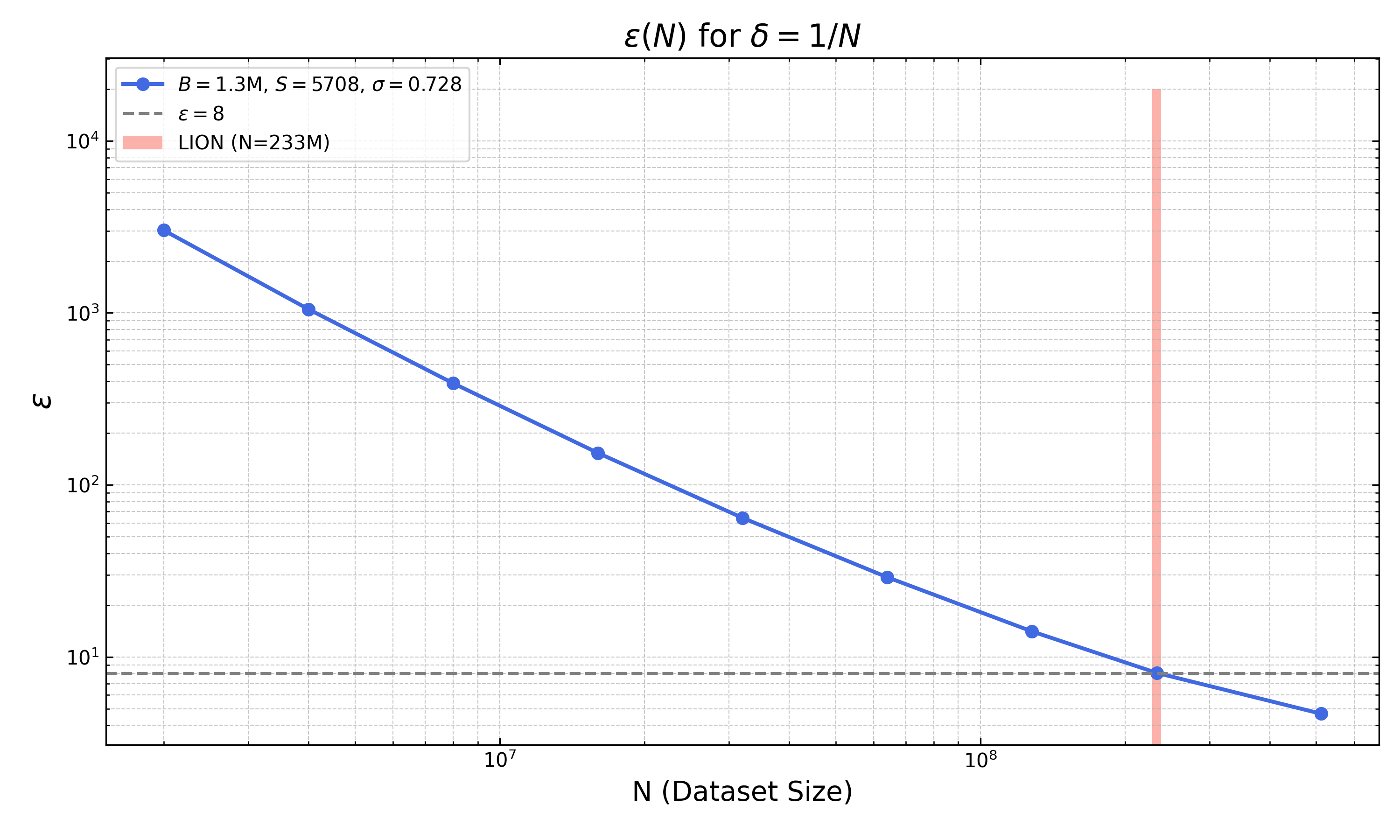}
\vspace{-0.1in}
    \caption{At fixed (B, $\sigma$, S), $\varepsilon$ drastically reduces with the dataset size.}
    \label{fig:eps_N}
\end{figure}

\begin{figure}[t!]
     \centering
\includegraphics[width=0.7\textwidth]{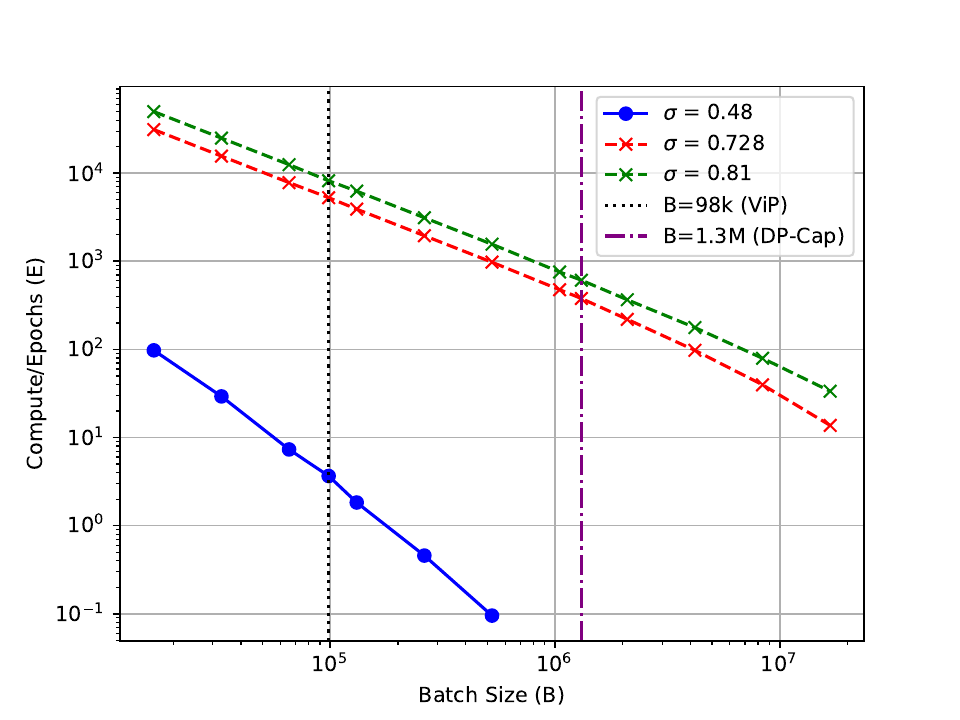}
    \vspace{-0.09in}
    \caption{All points correspond to $\varepsilon=8$ for a dataset of size $N=233$M.  At fixed $\varepsilon$ and $\sigma$, the number of epochs decreases as the batch size increases.}
    \label{fig:cap_privacy_parameters2}
\end{figure}

\textbf{Batch size and $\sigma$.} We wish to underscore the influence of batch size and $\sigma$ on both the computational budget and model performance. As highlighted in Section~\ref{subsec:effective_strategies_DPCAP}, for a given target $\varepsilon$, elevating $\sigma$ beyond 0.5 allows training for significantly more steps.
In Figure~\ref{fig:cap_privacy_parameters2}, the blue, orange and green lines show the batch size ($B$) vs. compute trade-off ($E$) at a given $\sigma$. The lines are monotonically decreasing with $B$, signifying that the number of epochs $E$ decreases when increasing $B$. When maintaining a fixed privacy budget $\varepsilon=8$, even a marginal increase in $\sigma$ from $0.48$ to $0.728$ (from blue to orange) translates to a remarkable increase ranging from 100 (for small batch sizes) to 100,000 (for very large batch sizes) times more gradient steps. Thus it is favorable to increase $\sigma$ and $B$ at the same time for better model convergence.

Meanwhile, doing so also incurs a higher computational cost: 
Under a 32-epoch budget on Dedup-LAION-233M with a batch size of $1.3$ million, we had to cut the red curve in Figure~\ref{fig:cap_privacy_parameters2}, with $\sigma=0.728$.
As outlined in Section~\ref{subsec:dp-cap-scaling}, with twice this budget, we could have raised $\sigma$ to 0.81 (green curve), with simulations indicating that this would have substantially improved performance.
Additionally, Section~\ref{subsec:effective_strategies_DPCAP} underscores that increasing the batch size is pivotal for achieving a high SNR while maintaining reasonable privacy guarantees. 
It is also crucial to note that at fixed $\varepsilon$, the compute budget is inversely proportional to the batch size. 
Therefore, increasing the batch size is beneficial for both SNR and computational efficiency. However, an excessively large batch size leads to fewer epochs and consequently a very limited number of training steps, which is detrimental to the training process (in addition to the difficulties of large batch training). 
For optimal privacy-utility-compute trade-off, a balance must be struck between computational resources, feasible batch size, and a reasonable number of training steps.

\subsection{Image Caption Examples}\label{subsec:appendix-image-caption}

In Figures \ref{fig:good_captions_random} and \ref{fig:good_captions_top10}, we show images from the MS-COCO 2017 test set and their corresponding captions generated by human annotator, Cap, and DP-Cap. Images in Figure \ref{fig:good_captions_random} are selected randomly, whereas images in Figure \ref{fig:good_captions_top10} are randomly selected from the top 10\% CIDEr score examples for DP-Cap. Qualitatively, the human-generated captions are more precise, whereas the captions generated by Cap and DP-Cap are more generic and sometimes contain factual errors.
This is to be expected since Cap and DP-Cap are trained on LAION with much noisier text description and were \emph{not} fine-tuned on MS-COCO. Nevertheless, DP-Cap still generates grammatically correct and (mostly) semantically coherent captions for unseen images.

\begin{figure}[p]
     \centering
\includegraphics[width=0.9\textwidth,height=0.75\textheight]{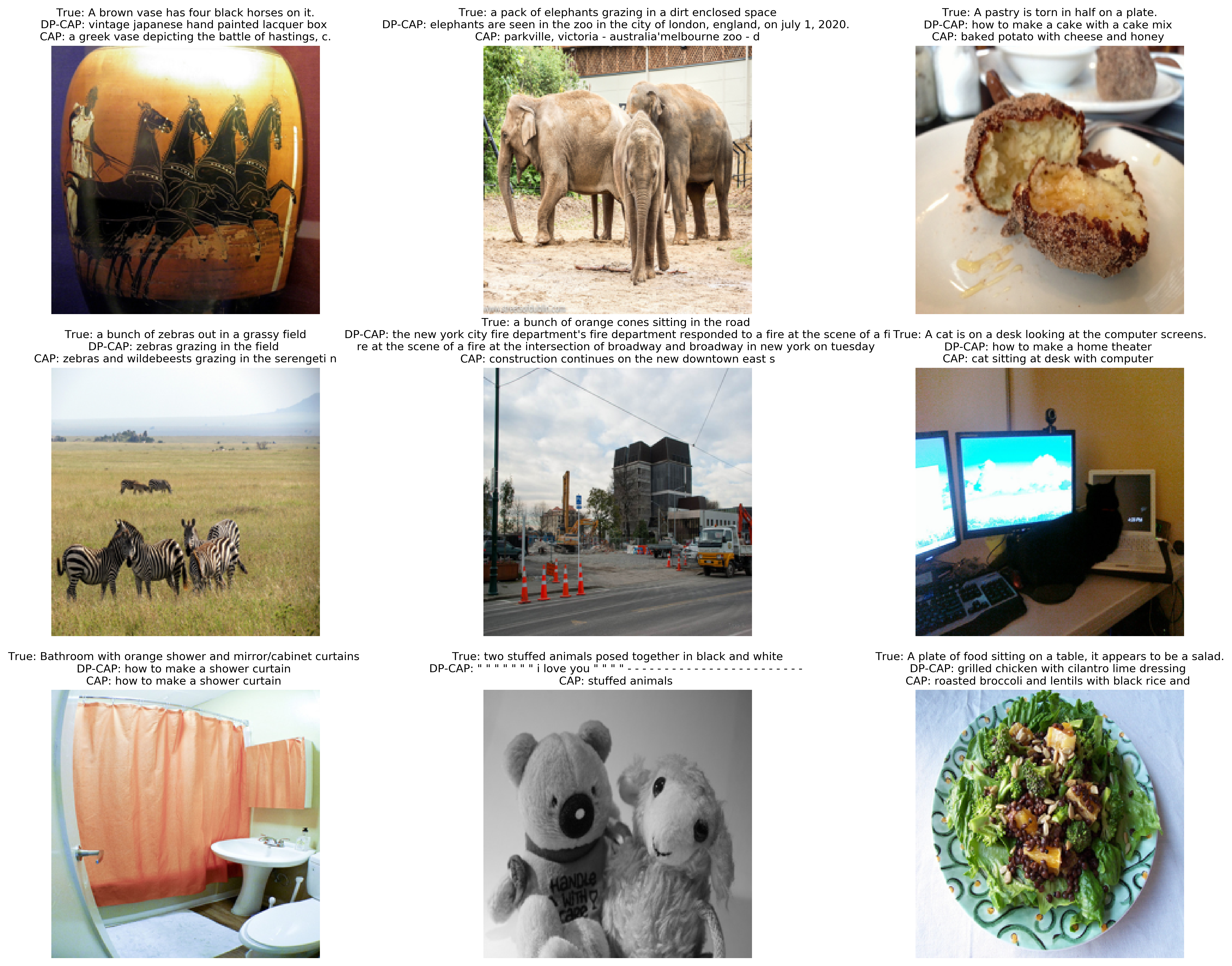}
     \vspace{-0.05in}
    \caption{Captions of randomly selected images from the MS-COCO 2017 test set.}
    \label{fig:good_captions_random}
    \vspace{-0.1in}
\end{figure}

\begin{figure}[p]
     \centering
\includegraphics[width=0.9\textwidth,height=0.75\textheight]{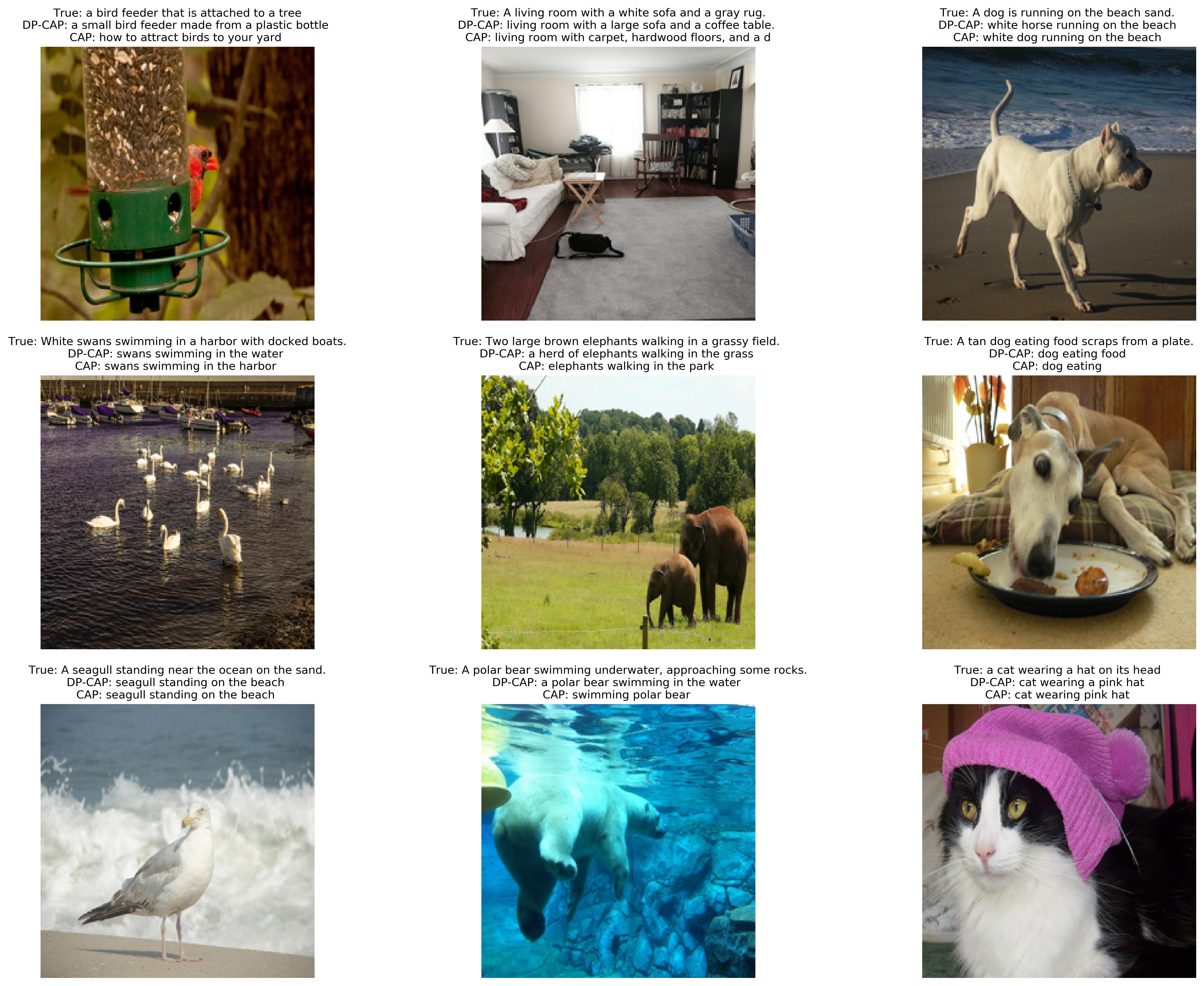}
     \vspace{-0.05in}
    \caption{Captions of images (randomly picked among the top 10\% CIDEr score of DP-Cap) from the MS-COCO 2017 test set.}
    \label{fig:good_captions_top10}
    \vspace{-0.1in}
\end{figure}